\documentclass[pdflatex,sn-mathphys,iicol]{sn-jnl}

\usepackage{
graphicx,
amsfonts,
textcomp,
program,
amssymb,
listings,
rotating,
booktabs,
algorithmicx,
mathrsfs,
url,
appendix,
etoolbox,
algpseudocode,
multirow,
algorithm,
amsmath,
hyperref,
manyfoot,
hhline,
pifont}

\usepackage{array}
\usepackage[export]{adjustbox}

\usepackage{multirow}
\usepackage{algpseudocode}
\usepackage{setspace}
\usepackage{subfigure}
\usepackage{bm}

\newcommand\shline{\specialrule{0.85pt}{0pt}{0pt}}

\jyear{2023}
\theoremstyle{thmstyleone}

\theoremstyle{thmstyletwo}

\theoremstyle{thmstylethree}

\begin{document}

\title[Article Title]{\vspace{-30pt}DiffLLE: Diffusion-guided Domain Calibration for Unsupervised Low-light Image Enhancement}

\author[1]{\fnm{Shuzhou} \sur{Yang}}\email{\scriptsize szyang@stu.pku.edu.cn }\equalcont{Equal contribution.}
\author[1]{\fnm{Xuanyu} \sur{Zhang}}\email{\scriptsize xuanyuzhang21@stu.pku.edu.cn }\equalcont{Equal contribution.}
\author[1]{\fnm{Yinhuai} \sur{Wang}}\email{\scriptsize yinhuai@stu.pku.edu.cn}
\author[1]{\fnm{Jiwen} \sur{Yu}}\email{\scriptsize yujiwen@stu.pku.edu.cn}
\author[1]{\fnm{Yuhan} \sur{Wang}}\email{\scriptsize yuhan.wang@stu.pku.edu.cn}
\author*[1]{\fnm{Jian} \sur{Zhang}}\email{zhangjian.sz@pku.edu.cn}
\affil[1]{\orgdiv{School of Electronic and
Computer Engineering, Peking University}, \orgaddress{\city{Shenzhen}, \country{China}}}

\abstract{Existing unsupervised low-light image enhancement methods lack enough effectiveness and generalization in practical applications. We suppose this is because of the absence of explicit supervision and the inherent gap between real-world scenarios and the training data domain. For example, low-light datasets are well-designed, but real-world night scenes are plagued with sophisticated interference such as noise, artifacts, and extreme lighting conditions. In this paper, we develop \textbf{Diff}usion-based domain calibration to realize more robust and effective unsupervised \textbf{L}ow-\textbf{L}ight \textbf{E}nhancement, called \textbf{DiffLLE}. Since the diffusion model performs impressive denoising capability and has been trained on massive clean images, we adopt it to bridge the gap between the real low-light domain and training degradation domain, while providing efficient priors of real-world content for unsupervised models. Specifically, we adopt a naive unsupervised enhancement algorithm to realize preliminary restoration and design two zero-shot plug-and-play modules based on diffusion model to improve generalization and effectiveness. The Diffusion-guided Degradation Calibration (DDC) module narrows the gap between real-world and training low-light degradation through diffusion-based domain calibration and a lightness enhancement curve, which makes the enhancement model perform robustly even in sophisticated wild degradation. Due to the limited enhancement effect of the unsupervised model, we further develop the Fine-grained Target domain Distillation (FTD) module to find a more visual-friendly solution space. It exploits the priors of the pre-trained diffusion model to generate pseudo-references, which shrinks the preliminary restored results from a coarse normal-light domain to a finer high-quality clean field, addressing the lack of strong explicit supervision for unsupervised methods. Benefiting from these, our approach even outperforms some supervised methods by using only a simple unsupervised baseline. Extensive experiments demonstrate the superior effectiveness of the proposed DiffLLE, especially in real-world dark scenarios.}


\maketitle

\begin{figure*}
    \centering
  \subfigure{
    \begin{minipage}[t]{0.14\linewidth}
	\centering
	\includegraphics[width=1.\linewidth]{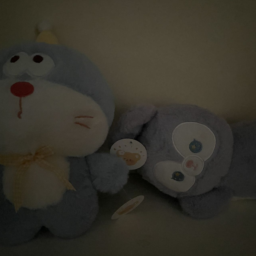}\\
        \centerline{Input}
        \includegraphics[width=1.\linewidth]{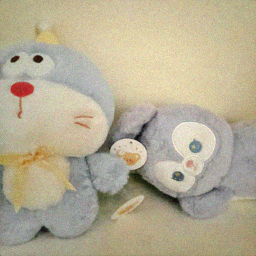}\\
        \centerline{KinD \cite{Zhang_2019_MM}}
    \end{minipage}%
}
    \subfigure{
    \begin{minipage}[t]{0.14\linewidth}
	\centering
	\includegraphics[width=1.\linewidth]{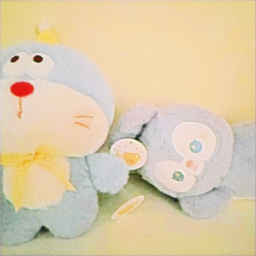}\\
        \centerline{SSIE \cite{SSIENet}}
        \includegraphics[width=1.\linewidth]{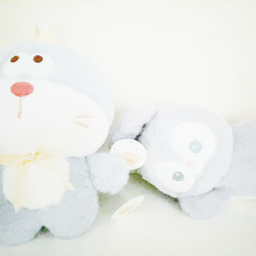}\\
        \centerline{SCI \cite{Ma_2022_CVPR}}
    \end{minipage}%
}
  \subfigure{
    \begin{minipage}[t]{0.14\linewidth}
	\centering
	\includegraphics[width=1.\linewidth]{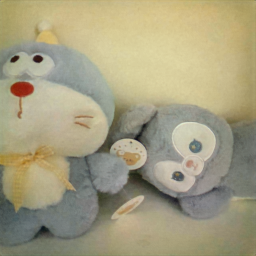}\\
        \centerline{EnGAN \cite{Jiang_2021_TIP}}
        \includegraphics[width=1.\linewidth]{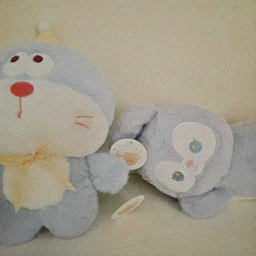}\\
        \centerline{URetinex \cite{Wu_2022_CVPR}}
    \end{minipage}%
}
  \subfigure{
    \begin{minipage}[t]{0.14\linewidth}
	\centering
	\includegraphics[width=1.\linewidth]{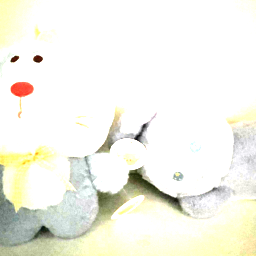}\\
        \centerline{RUAS \cite{Liu_2021_CVPR}}
        \includegraphics[width=1.\linewidth]{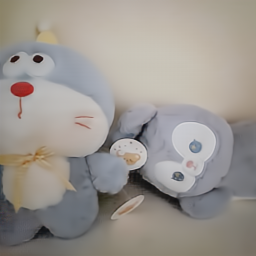}\\
        \centerline{Ours}
    \end{minipage}%
}
    \begin{minipage}{0.385\linewidth}
	\centering
        \vspace{3.5em}
	\includegraphics[width=1.\linewidth]{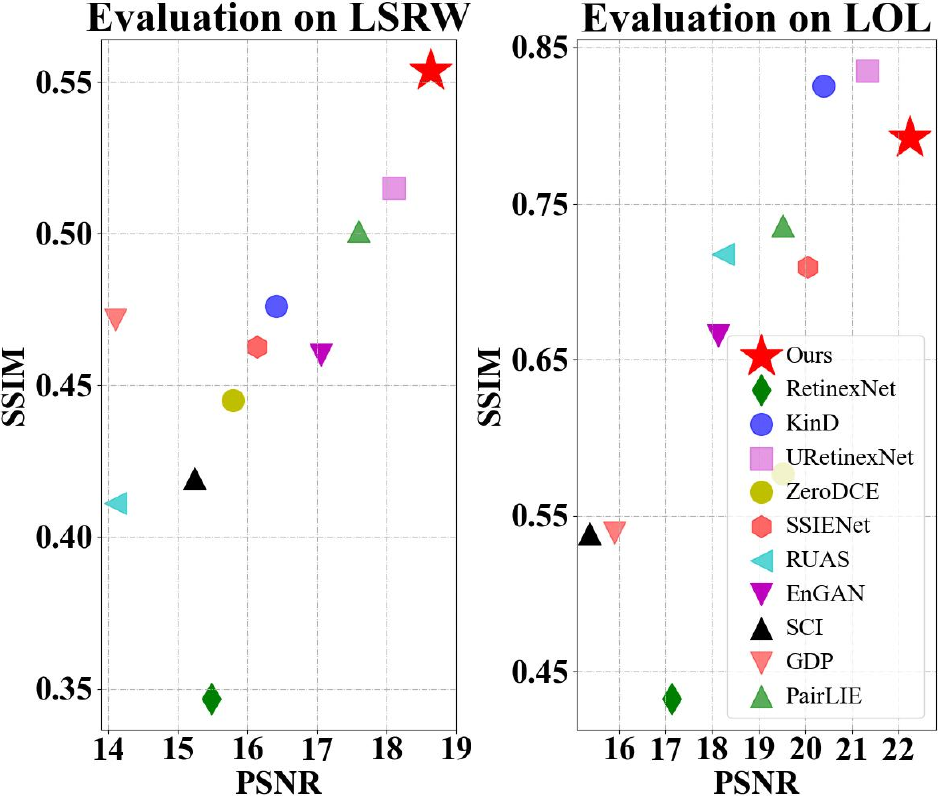}
    \end{minipage}
  \caption{Visual comparison of the proposed DiffLLE and other state-of-the-art methods on captured real-world low-light images show that our method not only removes natural noise, but also restores the most realistic color, while SSIE, EnGAN, and KinD fail to denoise and distort colors. URetinex loses color and remains noise. RUAS and SCI overexpose the image. The proposed DiffLLE achieves the best performance on both LSRW \cite{LSRW} and LOL \cite{LOL} benchmarks.}
  \label{fig:teaser}
\end{figure*}

\section{Introduction}
\label{sec:intro}
Low-light enhancement aims to ameliorate the quality and brightness of poorly illuminated images. Due to its ability to recover unseen regions and improve visual perception, over the past few years, ill-posed low-light enhancement has spawned many significant downstream applications, including object detection \cite{Liu_2016_ECCV, Liu_2022_TIP}, and semantic segmentation \cite{Islam_2020_IROS, Jiang_2022_TCSVT}.

Prolific methods have been proposed to recover low-light degraded scenarios \cite{Pisano_1998_JDI, Ng_2011_SIAM, Guo_2017_TIP, Guo_2020_CVPR, Jiang_2021_TIP, Xu_2022_CVPR, Yang_2023_arxiV, Fei_2023_CVPR, Wang_2023_arxiV, Wu_2023_CVPR, Jin_2023_CVPR, Fu_2023_CVPR1, Xu_2023_CVPR}, which can be divided into two categories: conventional physics-based methods and data-driven methods. The former formulates low-light degradation into a physical model, such as histogram equalization \cite{Pisano_1998_JDI} and Retinex theory \cite{Ng_2011_SIAM}. They reconstruct normal-light images by constructing the hand-crafted model priors but are limited in describing diverse low-light degradation factors. To automatically learn priors from large-scale data for better performance and speed, data-driven methods are developed. Algorithms based on supervised learning \cite{Zhang_2021_ICCV, Wu_2022_CVPR} learn an end-to-end black-box mapping based on paired low-light and high-light data. Although they have achieved certain successes, it is still difficult to capture the paired training data and enable the models to have enough generalization ability in unknown degradations. Unsupervised methods \cite{Jiang_2021_TIP} are proposed to ease the reliance on paired data. They tend to train a generative model and implicitly encode the conversion from the low-light domain to the normal-light domain \cite{Jiang_2021_TIP} or develop no-reference objective functions to train networks \cite{Guo_2020_CVPR}. However, due to the lack of sufficient constraints, such methods often result in poor reconstruction quality and are prone to excessive noise and artifacts with outrageous visual effects. To be noted, the existing low-light training data is limited and well-designed, \textit{i.e.}, without realistic degradations such as noise and JPEG compression, \textit{etc.} Therefore, existing data-driven approaches have two major drawbacks: \textbf{1)}, insufficient ability to model realistic complicated degradations; \textbf{2)}, deficient ability to model high-quality content in normal-light scenarios.

\let\thefootnote\relax\footnotetext{\textcolor{blue}{For reproducible research, the complete source code of the proposed DiffLLE will be made publicly available when this paper is accepted.}}

To address the above issues, we develop a \textbf{Diff}usion guided domain calibration framework for \textbf{L}ow-\textbf{L}ight \textbf{E}nhancement, dubbed \textbf{DiffLLE}, to solve the unknown out-of-domain degradation in real low-light scenes. Benefiting from the strong diffusion priors learned from large-scale datasets and the progressive sampling mechanism, we bridge the gap between the real-world low-light domain and training low-light domain, the coarse enhancement solution space and fine normal-light field. Specifically, a \textbf{D}iffusion-guided \textbf{D}egradation \textbf{C}alibration mechanism (DDC) is proposed to finetune the brightness and narrow the semantic distance between the degraded low-light images in the wild and the elaborate training data. Then, we employ a bi-directional unsupervised enhancement mapping for preliminary restoration, which is learned from massive unpaired low-light and normal-light images. Meanwhile, undesirable perspective effects such as noise and artifacts are eliminated by \textbf{F}ine-grained \textbf{T}arget domain \textbf{D}istillation (FTD), thus realizing an adaptive and robust low-light enhancement. As shown in Fig.~\ref{fig:teaser}, our method achieves the best results on both public benchmarks and the captured in-the-wild images with more authentic tones and less noise. In a nutshell, our contributions are summarized as follows.

\vspace{3pt}
\noindent \ding{113}~(1) By utilizing diffusion model to realize domain calibration, a plug-and-play unsupervised enhancement method, dubbed DiffLLE, is proposed to bridge the inherent gap between real degradation domain and high-quality normal-light domain.

\vspace{3pt}
\noindent \ding{113}~(2) A diffusion-guided degradation calibration process is designed, which adjusts extreme lightness, eliminates undesired interference and replenishes authentic details in a zero-shot manner.

\vspace{3pt}
\noindent \ding{113}~(3) A fine-grained target domain distillation mechanism is proposed to refine the coarse enhancement results and guide the optimization of the unsupervised method.

\vspace{3pt}
\noindent \ding{113}~(4) Extensive experiments demonstrate that our method outperforms all of the SOTA unsupervised methods and even some supervised methods in both high-quality benchmarks and real images in the wild.

\section{Related Work}
\subsection{Low-light Image Enhancement}
Various methods have been proposed to improve the visibility of low-light images. The model-based approaches are first widely adopted. Zia-ur~\textit{et al.} \cite{Zia-ur_2004_JEI} proposed the Retinex theory and decomposed a captured image into illumination and  reflectance. Guo~\textit{et al.} \cite{Guo_2017_TIP} constructed a lightness map for targeted enhancement. Ren~\textit{et al.} \cite{LECARM} selected a camera response model to adjust each pixel to ideal exposure. Hao~\textit{et al.} \cite{SDD} improved Retinex in a semi-decoupled way, which estimates illumination and depicts reflectance based on it. However, these methods require tedious hand-designed priors and are only applicable to specific scenarios.

In recent years, data-driven methods have attracted wide attention, benefiting from their ability of learning efficient priors from massive data automatically \cite{Cai_2018_TIP, Guo_2020_CVPR, Liu_2021_CVPR, Zhang_2021_ICCV, Xu_2022_CVPR, Wu_2022_CVPR, Yang_2023_arxiV, fei2023generative}. For example, Liu~\textit{et al.} \cite{Liu_2021_CVPR} characterized the intrinsic underexposed structure in low-light images and enhanced follow Retinex rule. Whilst Zhang~\textit{et al.} \cite{Zhang_2021_ICCV} captured the structural relationships across different patches in an image for illumination enhancement. Xu~\textit{et al.} \cite{Xu_2022_CVPR} exploited the inherent noise in dark regions as the enhancement guidance. Wu~\textit{et al.} \cite{Wu_2022_CVPR} designed three neural modules to recover images in three steps, which are responsible for initialization, optimization, and illumination adjustment. Yang~\textit{et al.} \cite{Yang_2023_arxiV} adopted neural representation to normalize degradation to ease enhancement difficulty. However, these methods lack sufficient effectiveness in real-world applications as the limited training data can never cover all possible dark degradation.

To this end, we exploit the prolific priors from diffusion model and develop the diffusion-based domain calibration. It extends the effect of a trained model to sophisticated real-world degradation robustly without any fine-tuning or retraining operations.

\subsection{Diffusion Models for Image Inverse Problem}
Recently, numerous denoising diffusion models \cite{Ho_2020_NIPS, Song_2021_ICLR, Nichol_2021_ICML, Dhariwal_2021_NIPS, meng2022sdedit} have been proposed to solve the image inverse problem and achieved impressive performance. Several approaches~\cite{Choi_2021_ICCV, Kawar_2022_NIPS, Wang_2023_ICLR} adopted diffusion models to modify clean intermediate results progressively. For instance, Choi~\textit{et al.}~\cite{Choi_2021_ICCV} used pre-defined filters to extract low-frequency signals and filter high-frequency signals and refine coarse reconstruction results at each step. Kawar~\textit{et al.}~\cite{Kawar_2022_NIPS} adopted singular value decomposition to perform the diffusion process in the spectral space and developed a posterior sampling strategy. Simultaneously, Wang~\textit{et al.} \cite{Wang_2023_ICLR} integrated range-null space decomposition and diffusion model prior to ensure data consistency with realistic details. Besides, Saharia~\textit{et al.} \cite{Saharia_2022_pami} directly cascaded low-resolution measurement and the latent code as input to train a conditional diffusion model for restoration. Liu~\textit{et al.} \cite{liu_2023_arxiv} established nonlinear diffusion bridges between the degraded and clean image distributions. Wang~\textit{et al.} \cite{Wang_2023_CVPR} employed the diffusion-based degradation remover to recover a coarse result and utilized a supervised pre-trained enhancement module to refine it with more authentic quality.

In this paper, we leverage the powerful generative capability of the diffusion model. It constrains unpredictable inputs to a specific degraded feature domain and distills a high-quality solution space. Compared with existing approaches, our method is plug-and-play, which can be directly applied to existing enhancement algorithms and achieve impressive performance gains.

\section{Preliminaries}
\label{sec:3}

\textbf{Diffusion process of DDIM \cite{Song_2021_ICLR}.}
Given a clean image $\mathbf{x}_0$ and random noise $\bm{\epsilon}\sim\mathcal{N}(0, \textbf{I})$, DDIM can be divided into two steps namely forward process and reverse sampling process. The forward process aims to add noise to the clean image $\mathbf{x}_0$. 
{\setlength\abovedisplayskip{0.1cm}
\setlength\belowdisplayskip{0.1cm}
\begin{equation}
\mathbf{x}_t=\sqrt{\overline{\alpha}_t} \mathbf{x}_{0}+\sqrt{1-\overline{\alpha}_t} \epsilon, \quad \epsilon \sim \mathcal{N}(\mathbf{0}, \mathbf{I}),
\end{equation}}where $\bar{\alpha}_t = \prod_{i=1}^t \alpha_t$ denotes the noise schedule and $\epsilon$ is a randomly sampled Gaussian noise variance. By utilizing a time-dependent pre-trained noise predictor $\boldsymbol{\epsilon}_{\boldsymbol{\theta}}(\cdot, t)$, the reversed sampling process aims to remove the noise step by step, which is defined as follows:
{\setlength\abovedisplayskip{0.1cm}
\setlength\belowdisplayskip{0.1cm}
\begin{equation}
    \mathbf{x}_s = \sqrt{\bar{\alpha}_s}\mathbf{f}_{\boldsymbol{\theta}}(\mathbf{x}_t, t) + \sqrt{1 - \bar{\alpha}_s - \sigma_t^2}\boldsymbol{\epsilon}_{\boldsymbol{\theta}}(\mathbf{x}_t, t) + \sigma_t\boldsymbol{\epsilon},
    \label{eq:ddim_sample}
\end{equation}
\begin{equation}
    with~~~~ \mathbf{f}_{\boldsymbol{\theta}}(\mathbf{x}_t, t) = (\mathbf{x}_t-\sqrt{1-\bar{\alpha}_t}\boldsymbol{\epsilon}_{\boldsymbol{\theta}}(\mathbf{x}_t, t))/ \sqrt{\bar{\alpha}_t},
\end{equation}}where $\mathbf{f}_{\boldsymbol{\theta}}(\cdot, t)$ denotes a denoising function. $\boldsymbol{\epsilon}_{\boldsymbol{\theta}}(\mathbf{x}_t, t)$ and $\boldsymbol{\epsilon}$ respectively denote the deterministic noise and random noise. Noting that it is not necessary to ensure the two steps in the DDIM sampling are adjacent (i.e., $t=s+1$). In fact, $s$ and $t$ can be any two steps that satisfy $s<t$, thus significantly accelerating the sampling process of the diffusion model. This process can also be equivalently depicted as solving an Ordinary Differential Equation (ODE)~\cite{Song_2021_ICLR}.

\begin{figure*}[t]
  \centering
    \includegraphics[width=0.95\linewidth]{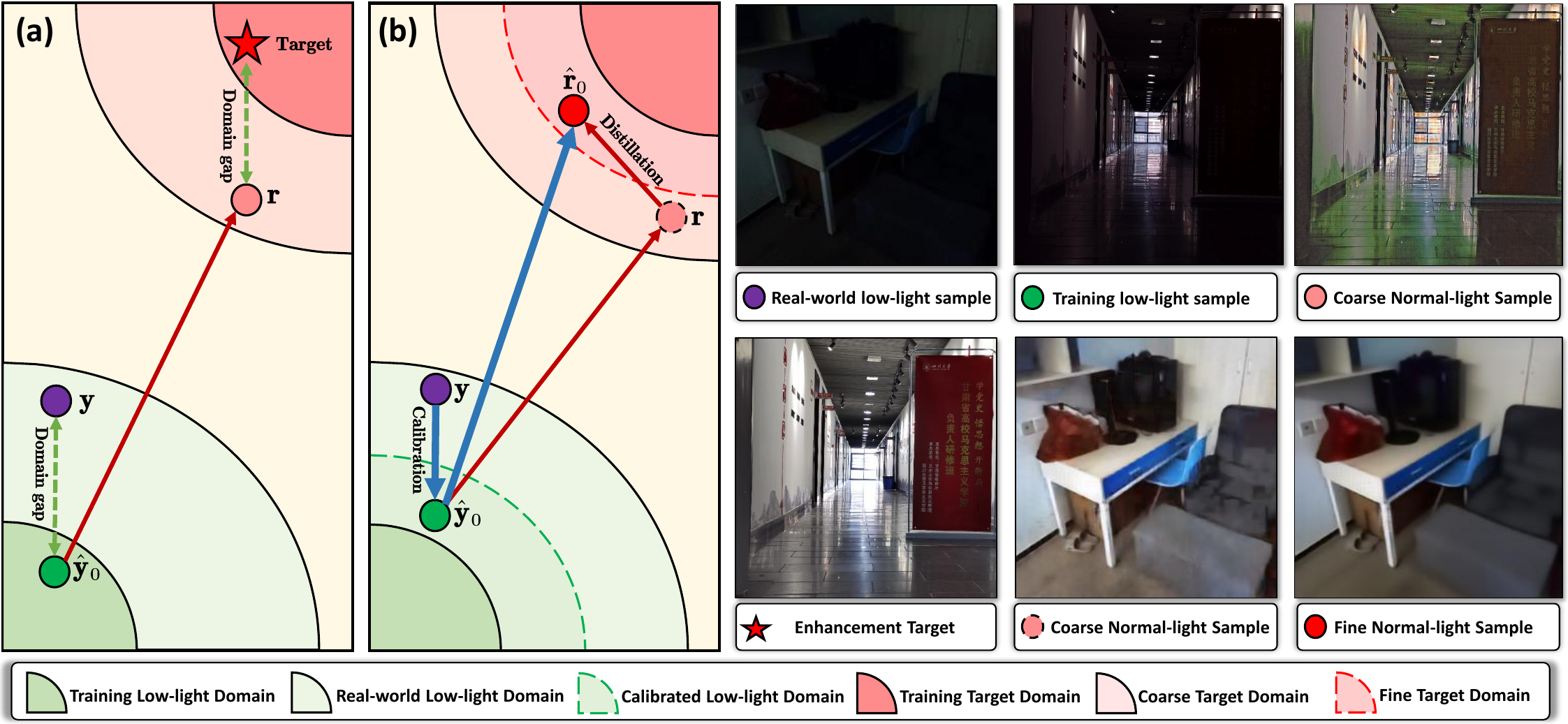}
  
  \caption{Visualization of different domains and domain conversions. (a) Workflow of the unsupervised learning-based methods, such as SSIENet \cite{SSIENet}, EnGAN \cite{Jiang_2021_TIP}, and CycleGAN \cite{CycleGAN}, \textit{etc.} It learns an enhancement mapping from the Training Low-light Domain (TLD) to the coarse normal-light domain, which cannot cover all possible dark conditions and converge to a fine clean target. (b) Workflow of the proposed DiffLLE. It calibrates in-the-wild low-light inputs to approximate TLD and recovers more high-quality results via domain distillation. The \textcolor{blue}{blue}, \textcolor[RGB]{112,173,71}{green}, and \textcolor{red}{red} respectively denote our inference process, the domain gap, and distillation operation.
  }
  
  \label{fig:fig2}
\end{figure*}

\section{Proposed Method}
\subsection{Diffusion-based Domain Calibration} 
To realize a realism-faithfulness trade-off, existing work~\cite{meng2022sdedit} has proved that image translation and domain transfer can be achieved by adding slight noise to the input and then removing noise step by step via the sampling process in Eq.~\ref{eq:ddim_sample} and a pre-trained denoiser. As the pre-trained denoiser has been trained on large clean high-quality datasets, the ``adding and removing'' process can translate a coarse source image (such as a stroke painting) to a realistic target image. Recently, \cite{gao2022back, yang2023synthesizing} have utilized diffusion-based domain transfer for image classification and restoration.

Particularly, targeted at low-light image enhancement, we surprisingly find that the degraded images processed by adding noise and removing it progressively tend to \textbf{calibrate the real-world complex low-light degradations and perform good properties with fewer artifacts and noise}. Benefiting from the powerful data-driven diffusion prior and the progressive denoising mechanism, it inherently narrows the domain gap between unsatisfactory intermediate results and the real-world clean images. We refer to the above process as ``\textbf{Diffusion-based Domain Calibration}''. Based on this process, we first adopt domain calibration for low-light images, which contains very complicated degradation, to reduce the difficulty of real-world low-light enhancement and improve our reconstruction quality. Besides, the priors from the diffusion model also benefit content modeling, improving the visual effect. We will elaborate them in more detail in Sec.~\ref{sec:4.3} and \ref{sec:4.4} respectively.

\subsection{Architecture of the Proposed DiffLLE}
\label{sec:4.2}
Our core motivation is to utilize diffusion-based domain calibration to enhance the perceptual quality and brightness of real-world low-light images. Based on an Unsupervised low-light image Enhancement Module (UEM), here we employ CycleGAN~\cite{CycleGAN} and you can also adopt other baselines such as EnGAN \cite{Jiang_2021_TIP}, \textit{etc.}, the proposed DiffLLE is mainly composed of two key components, namely Diffusion-guided Degradation Calibration (DDC) and Fine-grained Target domain Distillation (FTD). As illustrated in Fig.~\ref{fig:fig2} (b), given a degraded low-light image $\mathbf{y}$ in the real-world low-light domain (the light green region), the proposed DDC aims to narrow the discrepancy between the distribution of the data sample $\mathbf{y}$ and the elaborate Training Low-light Domain (TLD, the green region), which decreases enhancement difficulty for a trained model. We thus generate a pre-processed input $\hat{\mathbf{y}}_0$ close to the TLD. After enhancing it through UEM to obtain a coarse preliminary result, the proposed FTD enables the UEM to obtain better reconstruction quality, transferring the enhanced result from the coarse normal-light domain (the pink region) to the fine one (the red region). In conclusion, we obtain a clean normal-light result $\mathbf{r}$ as the following three steps.

\textbf{Setp 1: Training a UEM:} As shown in Fig.~\ref{fig:fig2} (a), based on the classical CycleGAN \cite{CycleGAN}, we first train a UEM on the elaborated unpaired datasets as follows: 
\begin{equation}
    \mathbf{r} = \mathcal{H}_{\text{UEM}}(\hat{\mathbf{y}}_0).
\end{equation}
Considering that UEM learns the mapping based on consistency constraint and generative adversarial supervision without any explicit supervision, its results are often accompanied by many uncontrollable artifacts and noise. That is, it can only recover coarse normal-light targets containing undesired artifacts and noise but fails to converge to a fine high-quality target domain. \textbf{Benefiting from the powerful generative capability of the pre-trained diffusion model, we alleviate this problem in this paper!}

\begin{figure*}[t]
  \centering
    \includegraphics[width=1.\linewidth]{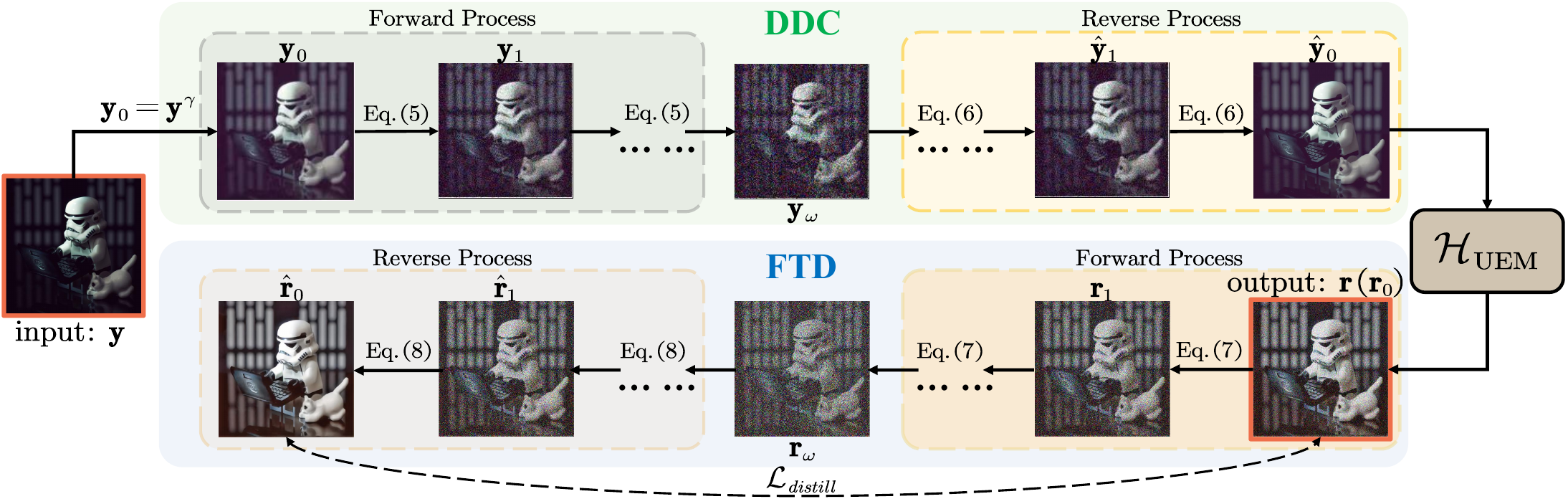}
  
  \caption{The architecture of DiffLLE. It consists of three components: Diffusion-guided Degradation Calibration (DDC), Unsupervised Enhancement Model (UEM), and Fine-grained Target domain Distillation (FTD). In inference, for the input $\mathbf{y}$, DDC uses a $\gamma$-curve to adjust its brightness ($\mathbf{y}_0$), here we set $\gamma=1.7$. Then a forward process generates a content-preserving latent code ($\mathbf{y}_{\omega}$). The following reverse process inverts $\mathbf{y}_{\omega}$ to a clean low-light intermediate image $\hat{\mathbf{y}}_0$. UEM enhances $\hat{\mathbf{y}}_0$ and produces the final result $\mathbf{r}$. During training, FTD is introduced to further refine $\mathbf{r}$, recovering a high-quality result $\hat{\mathbf{r}}_0$. We finetune UEM with $\hat{\mathbf{r}}_0$ for better effect.}
  \label{fig:DiffLLE}
\end{figure*}

\textbf{Step 2: Distilling the trained UEM:} As shown in the right part of Fig.~\ref{fig:fig2} (b), after training UEM, we utilize diffusion-based domain calibration to refine its coarse normal-light results. To eliminate the unsatisfactory visual effects, we first add a few steps of slight Gaussian noise to the coarse normal-light sample, converting it to the latent code of the diffusion model. Then we implement the inverse process through a pre-trained diffusion denoiser and obtain a refined result. Interestingly, using these refined results to fine-tune our previously trained UEM, we improve its enhancement quality. This distilling strategy is called Fine-grained Target domain Distillation (FTD). We report its detailed process in Algo.~\ref{alg:train} and will further introduce it in Sec.~\ref{sec:4.4}. Noting that although here we choose CycleGAN as UEM, in theory, any enhancement model can be adopted, which means that FTD is conducive to any unsupervised pipeline. In the last part of Sec.~\ref{sec:5.4}, we apply FTD to other enhancement methods and achieve obvious performance gains, which proves its plug-and-play property.

\textbf{Step 3: Calibrating the degradation domain:} As shown in the left region of Fig.~\ref{fig:fig2} (b), although the above model has been able to achieve impressive effectiveness on a well-designed dataset, real-world scenarios often encounter unknown degradation and extremely poor visibility. Therefore, directly adopting a trained network to real low-light scenes is difficult because of its limited generalization ability. To this end, we utilize a lightness enhancement curve to adjust the extreme lightness conditions. Due to its simple pixel-wise mapping, the inherent natural noise in dark regions also increases significantly with the emergence of content. We further employ diffusion-based domain calibration to alleviate unreasonable visual artifacts with an efficient diffusion model prior. As it has been pre-trained on a large dataset, the diffusion prior has a strong generative capability to narrow the domain gap between out-of-domain data and high-quality training input data, thus easing the difficulty of the enhancement task. This pre-process strategy is called Diffusion-guided Degradation Calibration (DDC). We report its detailed process in Algo.~\ref{alg:test2} and will further introduce it in Sec.~\ref{sec:4.3}. Noting that DDC is only required when enhancing out-of-domain data. For images from the widely used benchmark, we directly apply a fine-tuned UEM (as shown in Algo.~\ref{alg:test1}).

\begin{figure}
\begin{minipage}{.45\textwidth}
    \begin{algorithm}[H]
    \caption{Training DiffLLE on the Standard Datasets}
    \raggedright$\mathbf{Require:y}\thinspace(\rm{Input}), \omega\thinspace(hyper\mbox{-}parameter), $ \\
    $\qquad\qquad\boldsymbol{\Phi}\thinspace(\rm{Parameters\; of\; the\; Pretrained\; \mathcal{H}_{\text{UEM}}})$
    \label{alg:train}
    \begin{algorithmic}[1]
        \While {not converge}
        \State $\mathbf{r}_0 = \mathcal{H}_{\text{UEM}}(\mathbf{y})$
        \For{$t = 0, ..., \omega-1$}
            \State $\epsilon\sim\mathcal{N}(\mathbf{0},\mathbf{I})$
            \State update $\mathbf{r}_{t+1}$ via Eq.~(\ref{eq: rt+1})
        \EndFor
        \State $\hat{\mathbf{r}}_{\omega} = \mathbf{r}_{\omega}$
        \For{$t = \omega, ..., 1$}
            \State update $\hat{\mathbf{r}}_{t-1}$ via Eq.~(\ref{eq: rt-1})
        \EndFor
        \State $\boldsymbol{\Phi} \leftarrow \operatorname{argmin}_{\boldsymbol{\Phi}}\|\hat{\mathbf{r}}_{0}-\mathbf{r}_0\|_1$  \textcolor{blue}{\Comment{$\mathcal{L}_{distill}$}}
        \EndWhile
        \State \textbf{return}
    \end{algorithmic}
    \end{algorithm}
\end{minipage}

\begin{minipage}{.45\textwidth}
    \begin{algorithm}[H]
    \caption{Testing DiffLLE on the In-Domain Data}
    \label{alg:test1}
    \raggedright$\mathbf{Require:y}\thinspace(\rm{Input})$
    \begin{algorithmic}[1]
        \State $\mathbf{r}_0 = \mathcal{H}_{\text{UEM}}(\mathbf{y})$
        \State \textbf{return $\mathbf{r}_{0}$}
    \end{algorithmic}
    \end{algorithm}
\end{minipage}

\begin{minipage}{.45\textwidth}
    \begin{algorithm}[H]
    \caption{Testing DiffLLE on the Out-of-Domain Data}
    \label{alg:test2}
    \raggedright$\mathbf{Require:y}\thinspace(\rm{Input}), \gamma\;and\;\omega\thinspace(hyper\mbox{-}parameter)$
    \begin{algorithmic}[1]
        \State $\mathbf{y}_0 = \mathbf{y}^{\gamma}$
        \For{$t = 0, ..., \omega-1$}
            \State update $\mathbf{y}_{t+1}$ via Eq.~(\ref{eq:4})
        \EndFor
        \State $\hat{\mathbf{y}}_{\omega} = \mathbf{y}_{\omega}$
        \For{$t = \omega, ..., 1$}
            \State update $\hat{\mathbf{y}}_{t-1}$ via Eq.~(\ref{eq:5})
        \EndFor
        \State $\mathbf{r}_0 = \mathcal{H}_{\text{UEM}}(\hat{\mathbf{y}}_{0})$
        \State \textbf{return} $\mathbf{r}_{0}$
    \end{algorithmic}
    \end{algorithm}
\end{minipage}
\end{figure}

\subsection{Diffusion-guided Degradation Calibration}
\label{sec:4.3}
In this section, we present the proposed Diffusion-guided Degradation Calibration (DDC) strategy, designed to improve the generalization capability to real-world low-light scenarios.

\textbf{Motivation.} The trained UEM severely relies on elaborate training data. However, existing datasets cannot cover all degradations in real-world dark scenes. Fine-tuning or retraining is required when migrating the model to unseen scenes. To this end, we design DDC to calibrate the input data, narrowing the gap between the training low-light domain and out-domain data to achieve robust performance without additional training.

\textbf{Method.} As shown in the green part of Fig.~\ref{fig:DiffLLE}, DDC first adjusts the lightness of the input $\mathbf{y}$ through a lightness enhancement curve. Here we adopt the conventional $\gamma$-curve. It maps $\mathbf{y}$ to $\mathbf{y}_0$ at the pixel-level: $\mathbf{y}_0 = \mathbf{y}^{\gamma}$. As a too high $\gamma$ increases natural noise and a too low value cannot realize effective adjustment, we set $\gamma = 1.7$ for a trade-off. Afterward, to eliminate natural noise and artifacts, DDC conducts two steps: $\mathbf{Forward}$ and $\mathbf{Reverse}$. The former diffuses $\mathbf{y}_0$ to a content-preserving latent code $\mathbf{y}_{\omega}$, and the latter recovers it to $\hat{\mathbf{y}}_0$. Compared with input $\mathbf{y}$, $\hat{\mathbf{y}}_0$ contains more uniform lightness and much less interference, which is closer to the well-designed training data in feature space. We formulate the forward process in DDC as:
\begin{equation}
\begin{aligned}
\label{eq:4}
    & \mathbf{y}_{t+1}=\sqrt{\alpha_{t+1}} \mathbf{y}_{t}+\sqrt{1-\alpha_{t+1}} \epsilon, \quad \epsilon \sim \mathcal{N}(\mathbf{0}, \mathbf{I}), \\
    & t = 0, 1, ..., \omega-1,
\end{aligned}
\end{equation} where $\mathbf{y}_{t}$ is the current image state and $\mathbf{y}_{t+1}$ is the next, $\alpha_t$ is the coefficient of our noise schedule. This procedure is also presented in steps 2 to 5 of Algo.~\ref{alg:test2}. After obtaining the latent code $\mathbf{y}_{\omega}$, we adopt the pre-trained denoiser from DDIM, which has been trained on the ImageNet dataset, to recover the target image $\hat{\mathbf{y}}_0$. This reverse process is expressed as:
\begin{equation}
\begin{aligned}
\label{eq:5}
    & \hat{\mathbf{y}}_{t-1} = \sqrt{\bar{\alpha}_{t-1}}\mathbf{f}_{\boldsymbol{\theta}}(\hat{\mathbf{y}}_{t}, t) + \sqrt{1 - \bar{\alpha}_{t-1}}\boldsymbol{\epsilon}_{\boldsymbol{\theta}}(\hat{\mathbf{y}}_{t}, t), \\
    & t = \omega, \omega-1, ..., 1,
\end{aligned}
\end{equation} where $\mathbf{f}_{\boldsymbol{\theta}}(\cdot, t)$ represents the denoising function and $\boldsymbol{\epsilon}_{\boldsymbol{\theta}}(\hat{\mathbf{y}}_{t}, t)$ is the estimated noise. $\alpha_t$ is the predefined coefficient. In each step, the denoiser predicts noise $\boldsymbol{\epsilon}_{\boldsymbol{\theta}}(\hat{\mathbf{y}}_{t}, t)$ and recovers a clean image $\mathbf{f}_{\boldsymbol{\theta}}(\hat{\mathbf{y}}_{t}, t)$. Based on Eq.~\ref{eq:5}, the reverse process gradually produces the next state until $\hat{\mathbf{y}}_{0}$. Note that we set $\alpha_t = 1$ when $\textit{t}=1$, which means in the last step, we only denoise without adding noise, so as to get a clean result. Details are given in steps 6 to 9 of Algo.~\ref{alg:test2}. As the denoiser has been trained on massive well-designed data, in the reverse process, it is able to effectively eliminate noise and artifacts, producing images that are close to the well-designed training data domain. Benefiting from this, unlike other methods that require fine-tuning or retraining when migrating to unseen datasets, our method achieves robust and impressive performance through DDC without training, which is proved in Sec.~\ref{sec:5.3}.

\subsection{Fine-grained Target Domain Distillation}
\label{sec:4.4}
In this section, we illustrate the proposed Fine-grained Target domain Distillation (FTD). It refines the learned target domain of UEM. Noting that FTD is only used for the training purpose.

\textbf{Motivation.} To achieve more practical low-light enhancement, we train the model in an unsupervised manner, based on CycleGAN. However, it relies only on discriminator and consistency constraints, which are insufficient to provide sufficient supervision. As shown in the blue region of Fig.~\ref{fig:DiffLLE}, the enhanced result $\mathbf{r}$ still remains undesired noise. Hence, we develop FTD to refine the preliminary results and use them as pseudo-references to fine-tune UEM.

\textbf{Method.} Similar to DDC in Sec.~\ref{sec:4.3}, FTD also refines $\mathbf{r}$ to $\hat{\mathbf{r}}_0$ through the forward and reverse steps. As shown at the bottom of Fig.~\ref{fig:DiffLLE}, the forward process adds noise step by step to generate the latent code $\mathbf{r}_{\omega}$, and the reverse step reverses back to the clean domain and generates the target normal-light sample $\hat{\mathbf{r}}_0$. To illustrate the forward process more intuitively, we formulate it as:
\begin{equation}
\begin{aligned}
    & \mathbf{r}_{t+1} = \sqrt{\alpha_t} \mathbf{r}_{t}+\sqrt{1-\alpha_t} \epsilon,\quad \epsilon \sim \mathcal{N}(\mathbf{0}, \mathbf{I}), \\ 
    & t = 0, 1, ..., \omega-1.
\end{aligned}
\label{eq: rt+1}
\end{equation} It is similar to Eq.~\ref{eq:4}, but the input is the coarse enhanced result $\mathbf{r}$ rather than low-light image $\mathbf{y}_0$. To unify expression, here we express $\mathbf{r}$ as $\mathbf{r}_0$, 0 means its state number in the forward chain. After obtaining latent code $\mathbf{r}_{\omega}$, the same pre-trained denoiser is utilized to recover, expressed as:
\begin{equation}
\begin{aligned}
    & \hat{\mathbf{r}}_{t-1} = \sqrt{\alpha_t}\mathbf{f}_{\boldsymbol{\theta}}(\hat{\mathbf{r}}_{t}, t) + \sqrt{1 - \alpha_t}\boldsymbol{\epsilon}_{\boldsymbol{\theta}}(\hat{\mathbf{r}}_{t}, t), \\
    & t = \omega, \omega-1, ..., 1.
\end{aligned}
\label{eq: rt-1}
\end{equation} It is similar to Eq.~\ref{eq:5} but the input is another latent code $\mathbf{r}_{\omega}$. Noting that, benefiting from the efficient priors of the diffusion model, this denoiser even adjusts tone to be more realistic, and complements satisfactory details. Thus, $\hat{\mathbf{r}}_{0}$ exhibits both visual-friendly brightness and texture. However, recovering from $\mathbf{r}_0$ to $\hat{\mathbf{r}}_{0}$ requires many iterations, and calling the diffusion model is also time-consuming. To this end, we only generate $\hat{\mathbf{r}}_{0}$ in the training stage and use it to fine-tune UEM. Namely, distilling UEM with priors of the pre-trained diffusion model. So that in inference, we can only use UEM to get $\hat{\mathbf{r}}_{0}$-like results. We develop a distillation objective function as:
\begin{equation}
    \mathcal{L}_{distill} = \|\hat{\mathbf{r}}_{0}-\mathbf{r}_0\|_1.
\end{equation}
Noting that FTD can be also directly applied to other unsupervised enhancement methods and achieves performance gains. We utilize FTD for other pipelines and provide results in Sec.~\ref{sec:5.4} to prove it. The detailed algorithm of FTD is also presented in Algo.~\ref{alg:train}.

\begin{table*}[t]
    \centering
    \caption{\label{tab:compare} Quantitative comparison between our method and different state-of-the-art methods. The best and the second-best results are highlighted in \textbf{\color{red}red} and \textbf{\color{blue}blue} respectively.}

    \resizebox{\linewidth}{!}{
    \begin{tabular}{c|c|c c c|c c c c c c c|c}
        \shline
	\multirow{2}{*}{Datasets} & \multirow{2}{*}{Metrics} & \multicolumn{3}{c|}{\cellcolor{gray!40}{Supervised Learning Methods}} & \multicolumn{8}{c}{\cellcolor{gray!40}{Unsupervised Learning Methods}} \\
	& & \cellcolor{gray!40}RetinexNet & \cellcolor{gray!40}KinD & \cellcolor{gray!40}URetinexNet & \cellcolor{gray!40}ZeroDCE & \cellcolor{gray!40}SSIENet & \cellcolor{gray!40}RUAS & \cellcolor{gray!40}EnGAN & \cellcolor{gray!40}SCI & \cellcolor{gray!40}PairLIE & \cellcolor{gray!40}CLIP-LIT & \cellcolor{gray!40}Ours \\ \shline
	\multirow{4}{*}{LOL \cite{LOL}} & PSNR $\uparrow$ & 17.13 & 20.38 & \textbf{\color{blue}21.33} & 16.31 & 20.04 & 18.23 & 18.13 & 15.38 & 19.51 & 12.39 & \textbf{\color{red}22.24} \\
	& SSIM $\uparrow$ & 0.4329 & \textbf{\color{blue}0.8254} & \textbf{\color{red}0.8350} & 0.5767 & 0.7093 & 0.7174 & 0.6655 & 0.5384 & 0.7358 & 0.4934 & 0.7923 \\ 
        & NIQE $\downarrow$ & 9.69 & 5.15 & 3.51 & 8.10 & 3.71 & \textbf{\color{blue}3.27} & 4.84 & 8.37 & 4.71 & 8.79 & \textbf{\color{red}3.09} \\
        & LOE $\downarrow$ & 609.9 & 445.1 & \textbf{\color{blue}204.9} & 240.3 & 286.7 & 227.7 & 405.1 & 280.9 & 232.1 & 355.5 & \textbf{\color{red}202.4} \\ \midrule
	\multirow{4}{*}{LSRW \cite{LSRW}} & PSNR $\uparrow$ & 15.48 & 16.41 & \textbf{\color{blue}18.10} & 15.80 & 16.14 & 14.11 & 17.06 & 15.24 & 17.60 & 13.48 & \textbf{\color{red}18.63} \\
	& SSIM $\uparrow$ & 0.3468 & 0.4760 & \textbf{\color{blue}0.5149} & 0.4450 & 0.4627 & 0.4112 & 0.4601 & 0.4192 & 0.5009 & 0.3962 & \textbf{\color{red}0.5536} \\
        & NIQE $\downarrow$ & 4.31 & \textbf{\color{blue}2.97} & 3.27 & 3.34 & 3.64 & 3.68 & 3.03 & 3.66 & 3.30 & 3.52 & \textbf{\color{red}2.79} \\
		& LOE $\downarrow$ & 535.6 & 255.4 & 202.4 & 216.0 & \textbf{\color{red}196.0} & \textbf{\color{blue}198.9} & 385.1 & 234.6 & 267.5 & 290.3 & 201.3 \\ \shline
    \end{tabular}
}
\end{table*}

\section{Experiments}
In this section, we first introduce the implementation details of our approach in Sec.~\ref{sec:5.1}. Then we conduct comparative experiments between DiffLLE and other State-of-the-Art (SOTA) methods in Sec.~\ref{sec:5.2}. To prove the effectiveness of the proposed plug-and-play components (\textit{i.e.}, DDC and FTD mentioned in Sec.~\ref{sec:4.3} and Sec.~\ref{sec:4.4}), ablation analyses are further provided in Sec.~\ref{sec:5.3}. All experiments are conducted on a single NVIDIA 3090 GPU and implemented based on PyTorch.

\subsection{Implement Details}
\label{sec:5.1}
We adopt the architecture of ResNet \cite{He_2016_CVPR} as the enhancement module, which contains 9 residual blocks. Following \cite{CycleGAN}, we employ patchGAN \cite{Isola_2017_CVPR} to construct the discriminator. It outputs a binary map instead of a value. During training, a CycleGAN is pre-trained firstly. Then we only employ its trained enhancement module as the UEM for following fine-tuning.

\textbf{Parameter Settings.} For \textbf{training} CycleGAN, we utilize Adam optimizer \cite{Adam} and set its hyper-parameters $\beta_1$ = 0.5, $\beta_2$ = 0.999, and $\epsilon$ = 10$^{-8}$. We train CycleGAN with 200 epochs, initialize the learning rate to 2$\times$10$^{-4}$ and decay it linearly to 0 in the last 100 epochs. The batch size is set to 1 and the patch size is resized to 256$\times$256 in the concern of efficiency. For \textbf{finetuning} the UEM, we simply adopt the Adam optimizer with default hyperparameters. We finetune the module for 100 epochs and the learning rate varies from 1$\times$10$^{-5}$ to 1$\times$10$^{-8}$ with the Cosine Annealing strategy. The batch size is 16 and the patch size is still 256$\times$256. In our experiments, the total number of DDIM iteration steps is set to 100. We select the final 3 steps to apply the noise addition and removal process. The ablation study on the iteration number is given in Sec.~\ref{sec:5.4}. We utilize the pre-trained diffusion model in ImageNet~\cite{dhariwal2021diffusion}.

\begin{figure*}[t]
  \centering
    \subfigure{
    \begin{minipage}[t]{0.15\linewidth}
	\centering
	\includegraphics[width=1.03\linewidth]{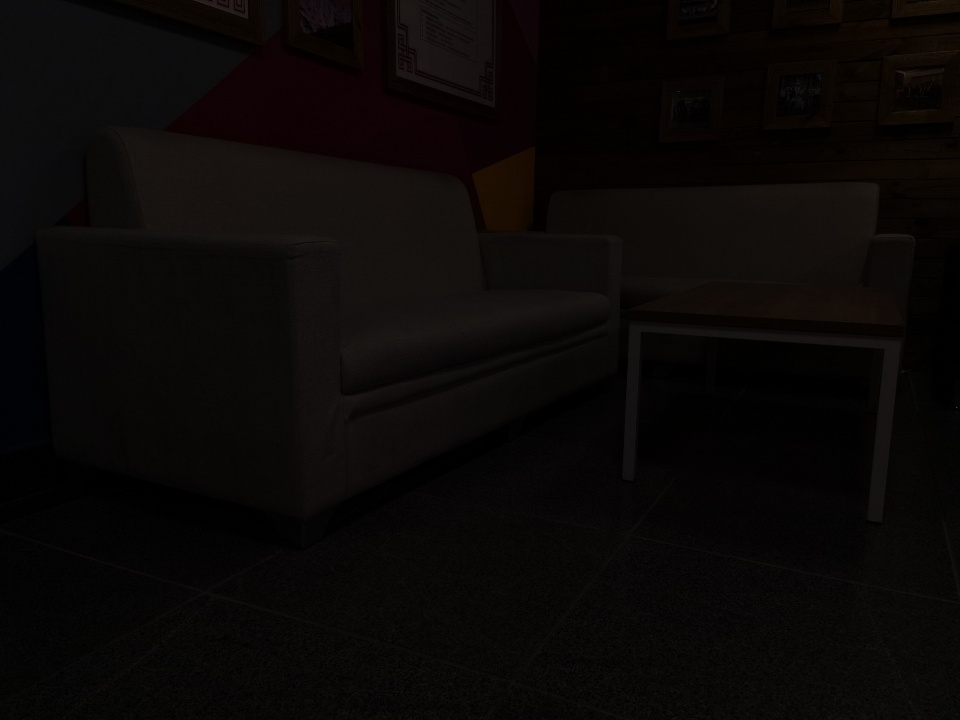}\\
        \centerline{\scriptsize Input(9.47)}
	\includegraphics[width=1.03\linewidth]{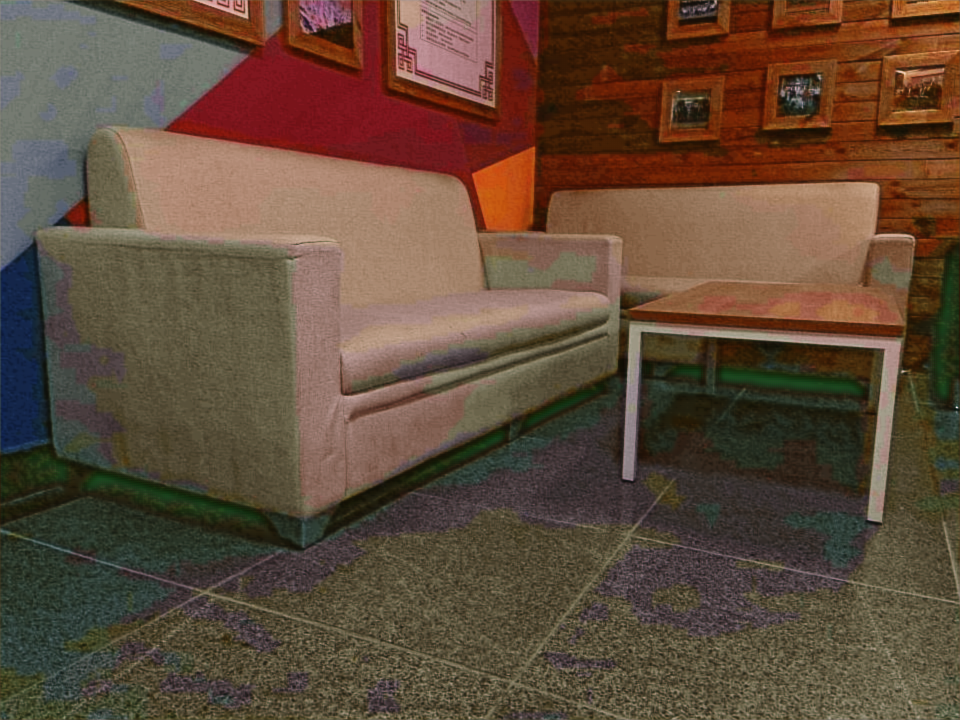}\\
        \centerline{\scriptsize SSIENet(18.78)}
    \includegraphics[width=1.03\linewidth]{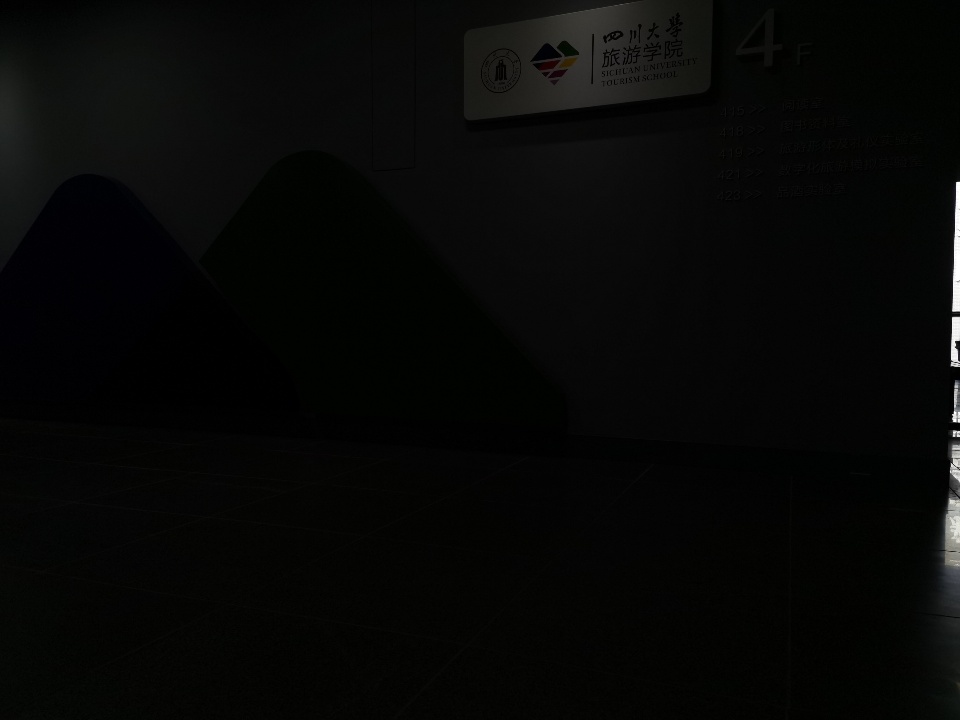}\\
        \centerline{\scriptsize Input(11.10)}
	\includegraphics[width=1.03\linewidth]{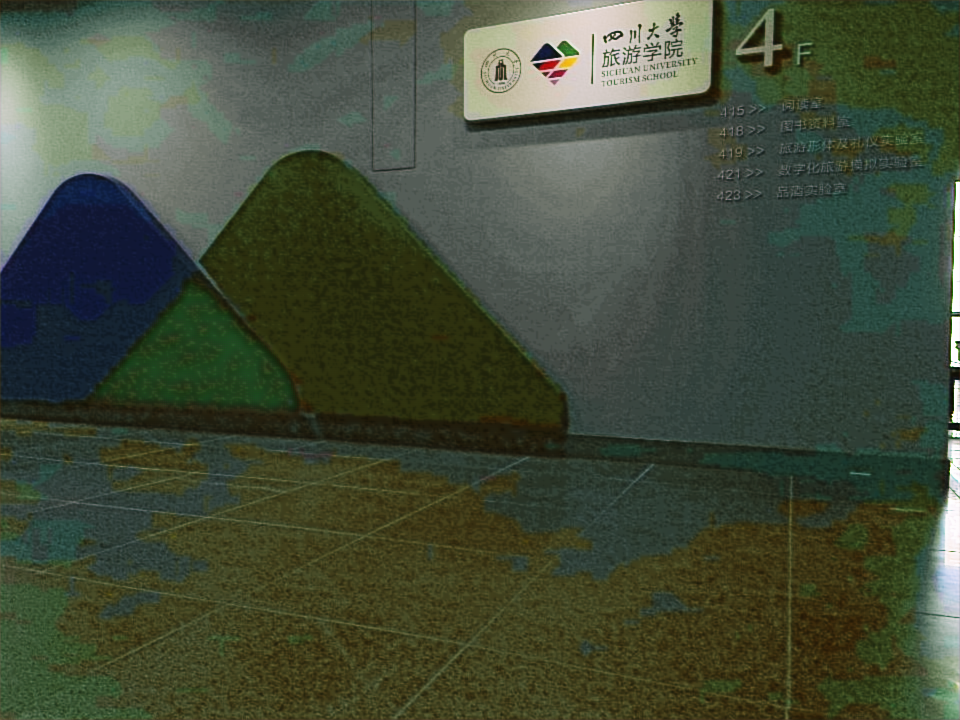}\\
        \centerline{\scriptsize SSIENet(22.20)}
    \end{minipage}%
}
  \subfigure{
    \begin{minipage}[t]{0.15\linewidth}
	\centering
	\includegraphics[width=1.03\linewidth]{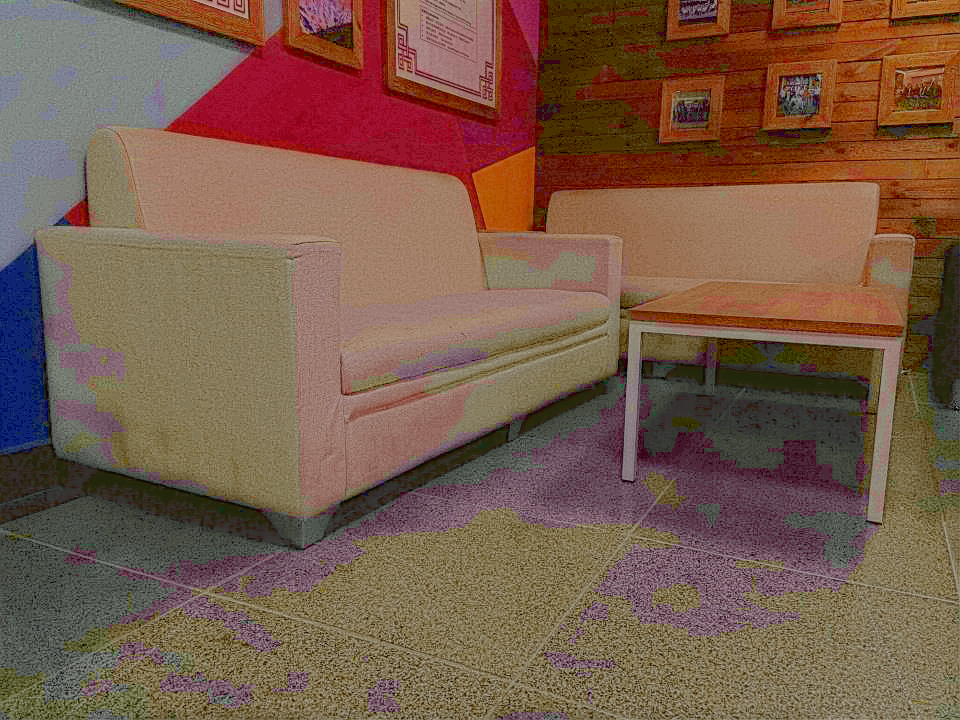}\\
        \centerline{\scriptsize RetinexNet(17.35)}
	\includegraphics[width=1.03\linewidth]{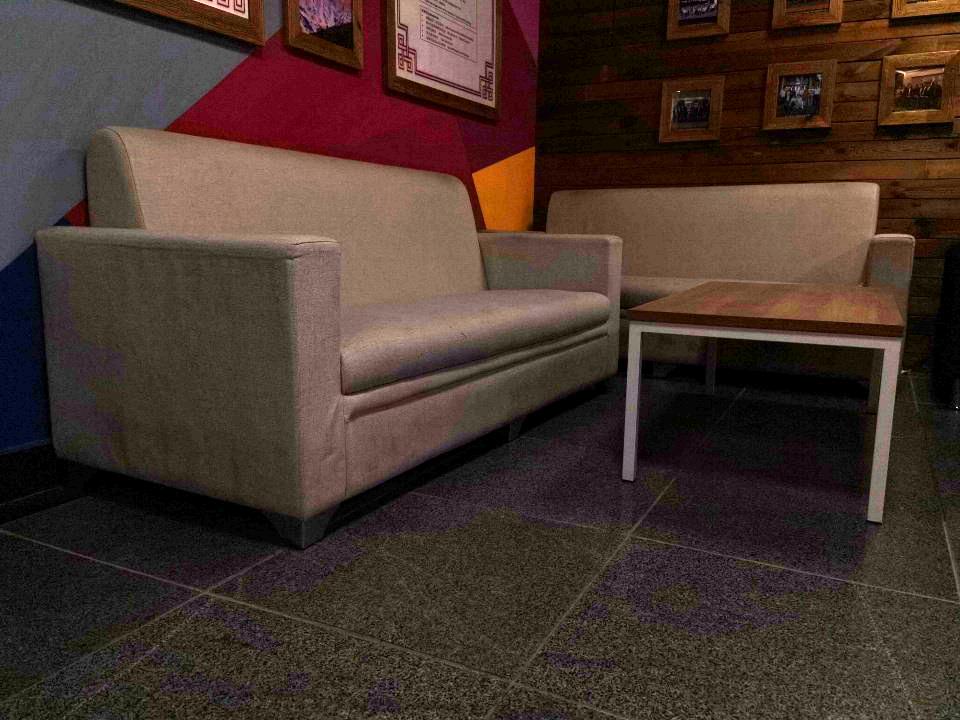}\\
        \centerline{\scriptsize RUAS(16.13)}
    \includegraphics[width=1.03\linewidth]{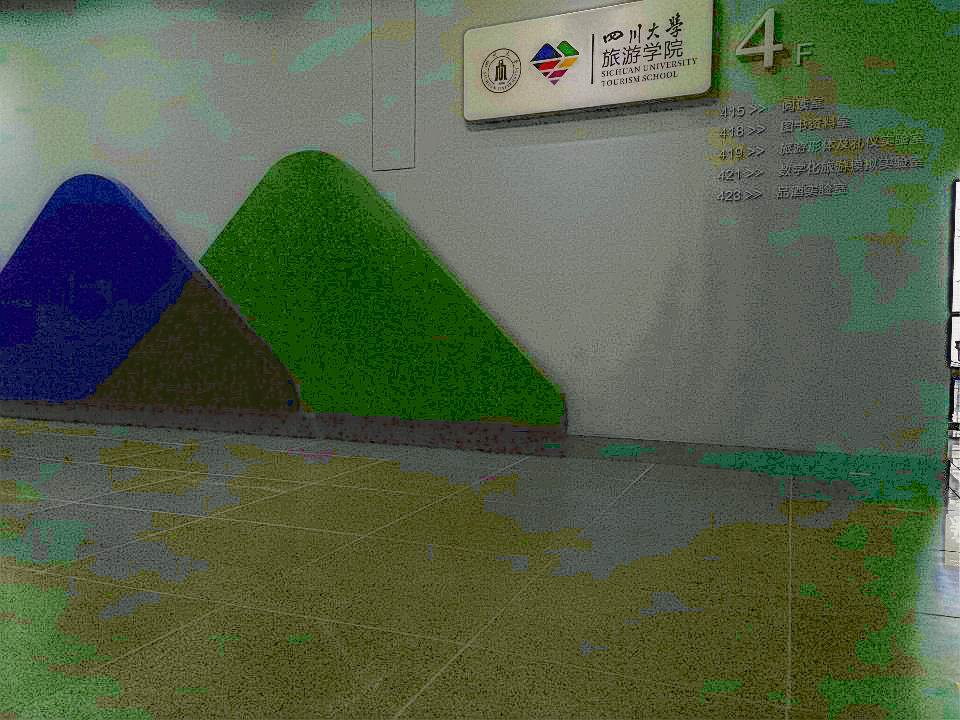}\\
        \centerline{\scriptsize RetinexNet(18.77)}
	\includegraphics[width=1.03\linewidth]{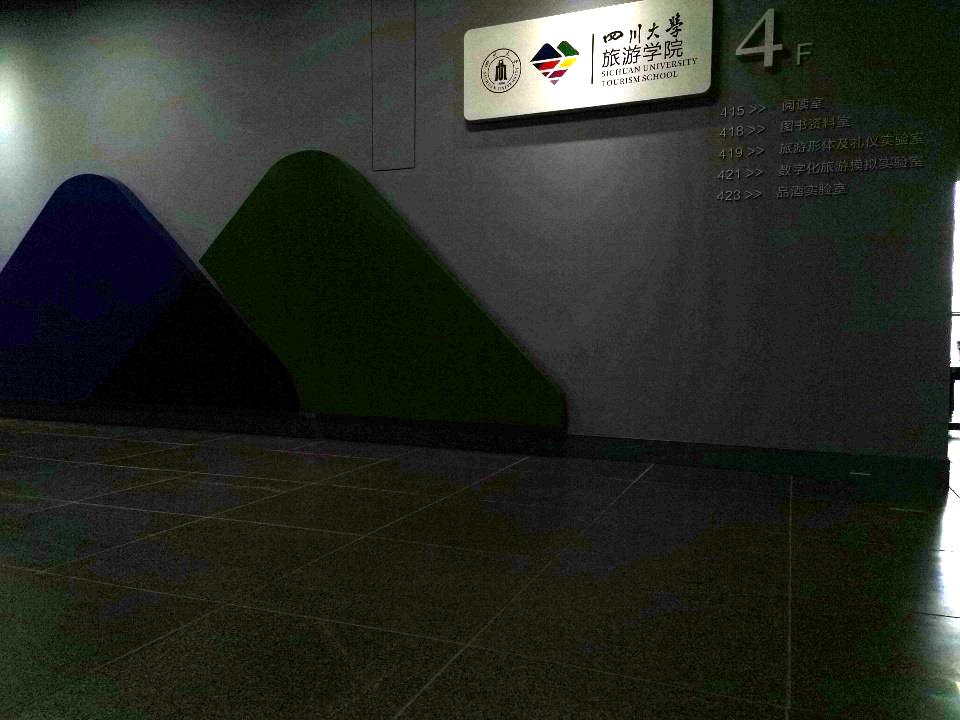}\\
        \centerline{\scriptsize RUAS(16.81)}
    \end{minipage}%
}
  \subfigure{
    \begin{minipage}[t]{0.15\linewidth}
	\centering
	\includegraphics[width=1.03\linewidth]{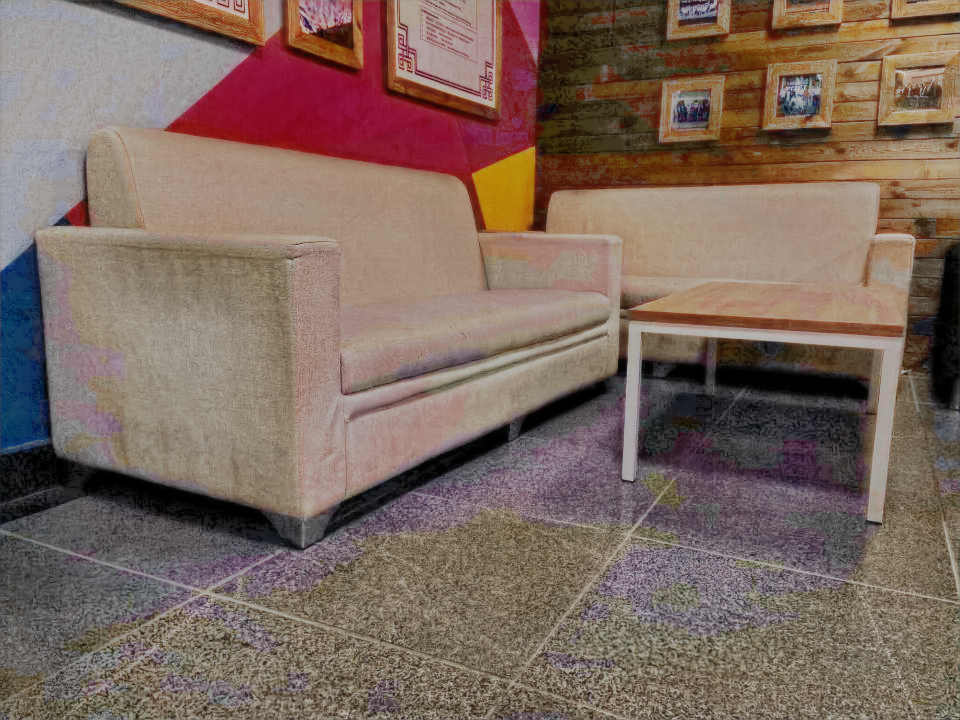}\\
        \centerline{\scriptsize KinD(16.58)}
	\includegraphics[width=1.03\linewidth]{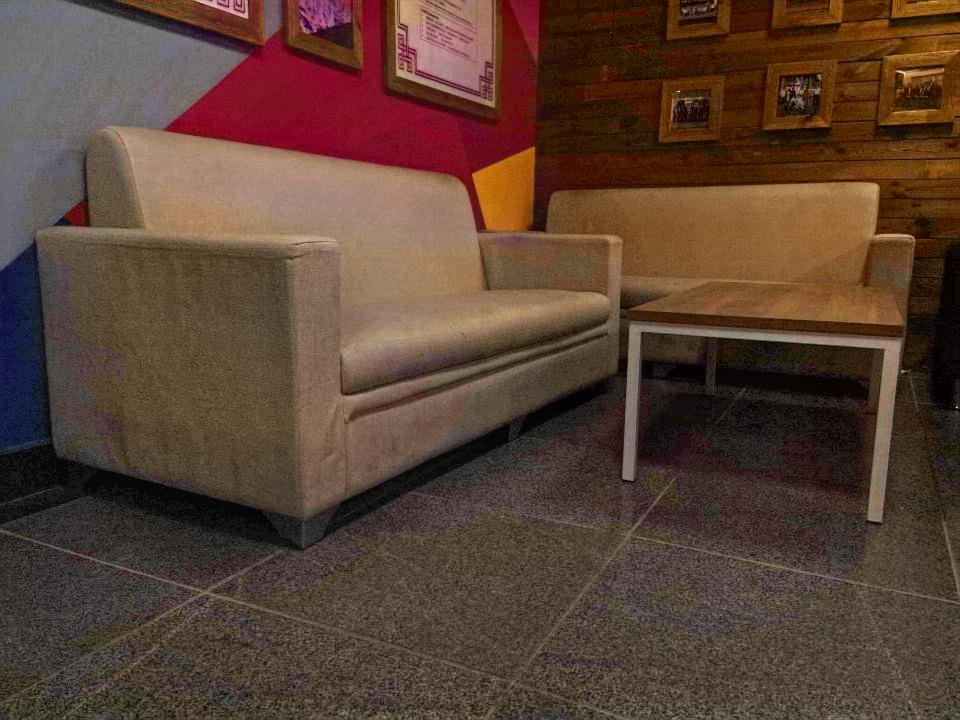}\\
        \centerline{\scriptsize EnGAN(17.91)}
    \includegraphics[width=1.03\linewidth]{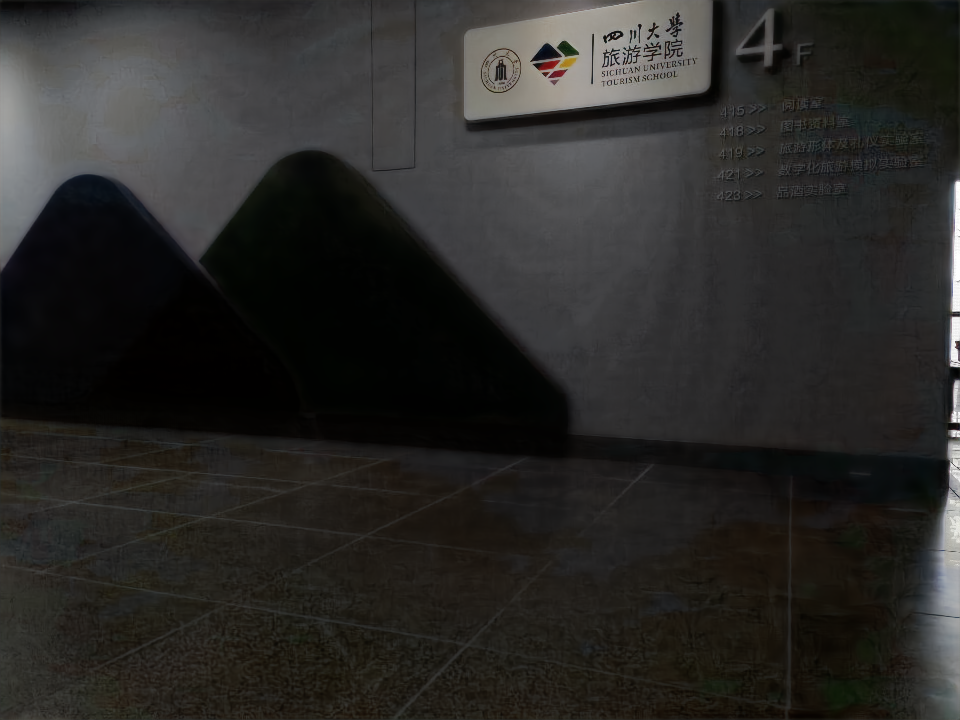}\\
        \centerline{\scriptsize KinD(17.45)}
	\includegraphics[width=1.03\linewidth]{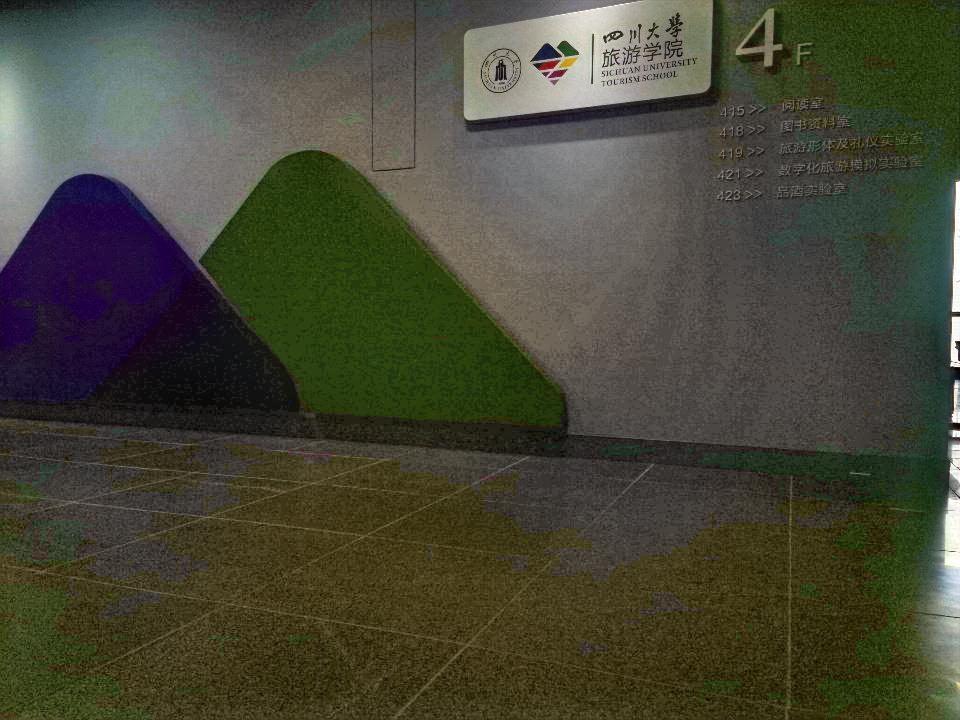}\\
        \centerline{\scriptsize EnGAN(21.85)}
    \end{minipage}%
}
  \subfigure{
    \begin{minipage}[t]{0.15\linewidth}
	\centering
	\includegraphics[width=1.03\linewidth]{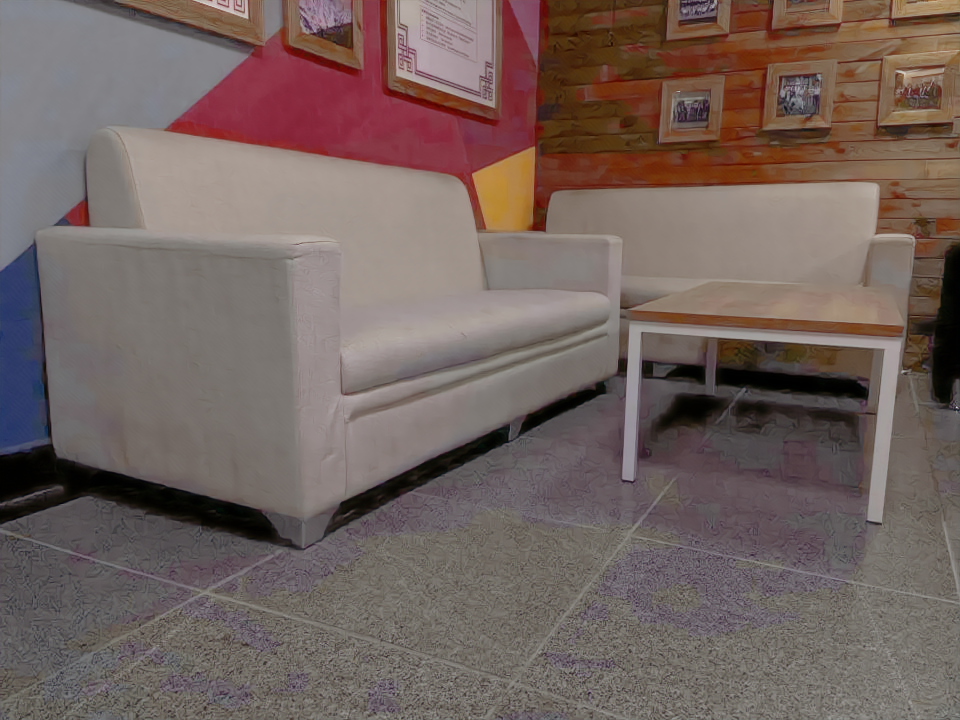}\\
        \centerline{\scriptsize URetinex(18.69)}
	\includegraphics[width=1.03\linewidth]{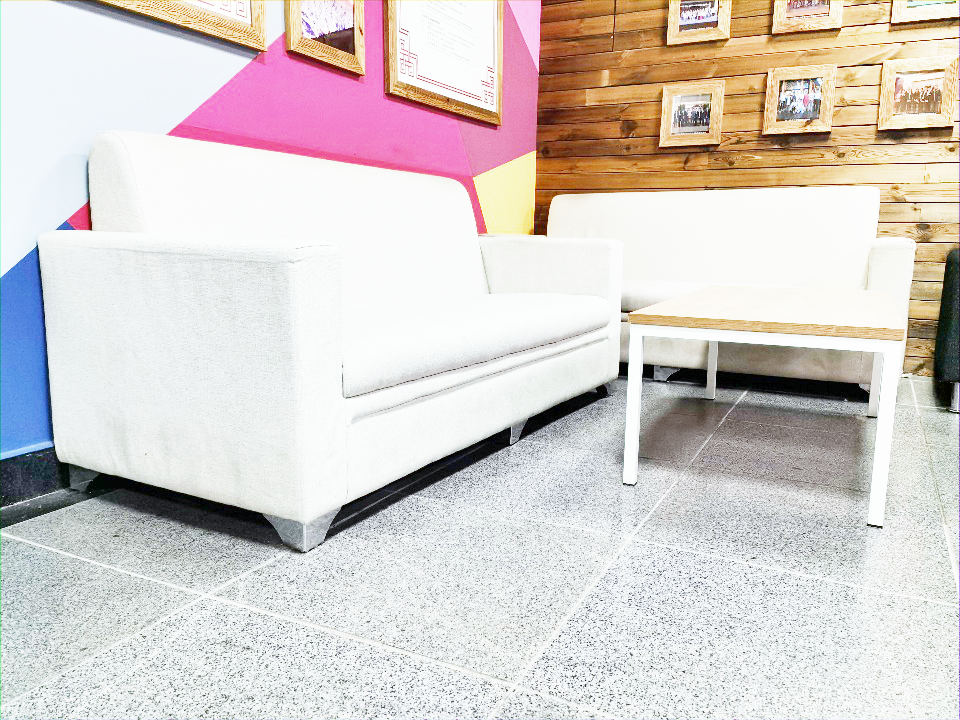}\\
        \centerline{\scriptsize SCI(5.99)}
	\includegraphics[width=1.03\linewidth]{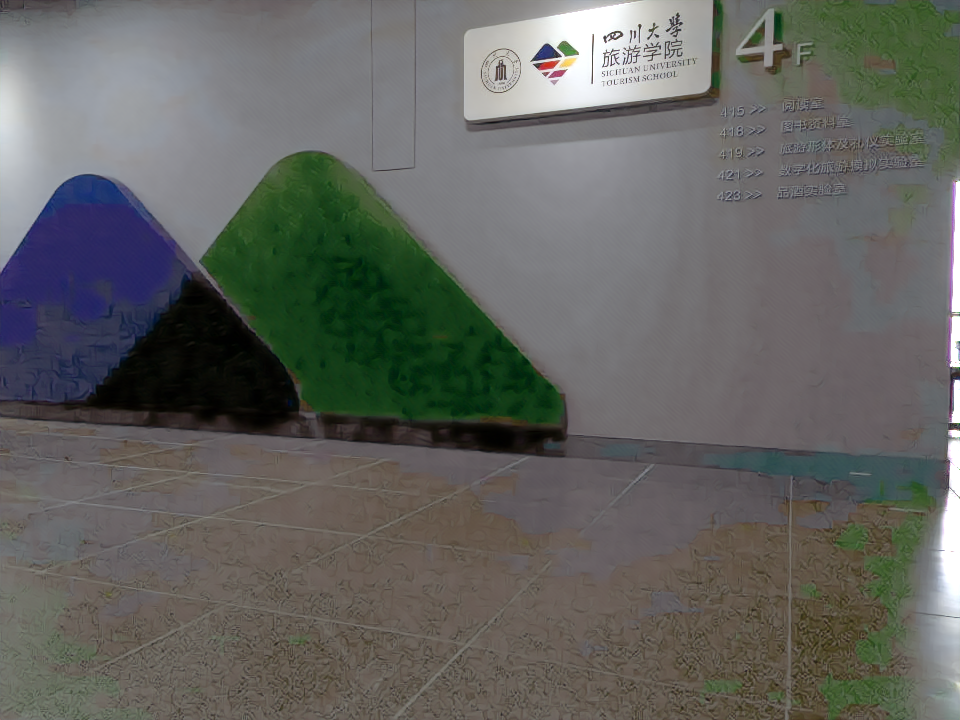}\\
        \centerline{\scriptsize URetinex(17.78)}
	\includegraphics[width=1.03\linewidth]{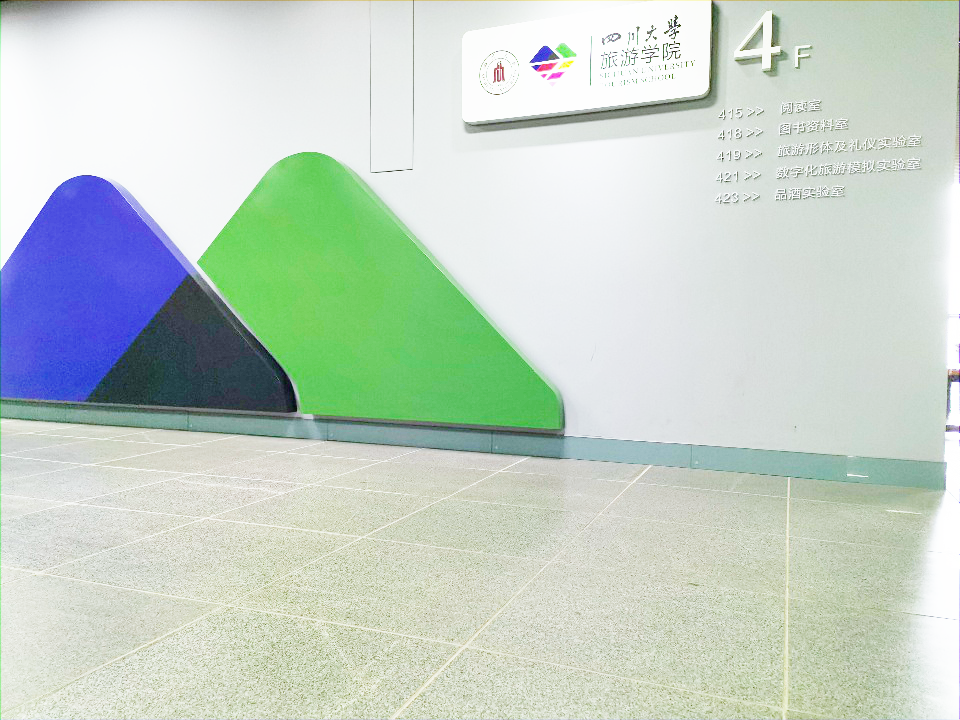}\\
        \centerline{\scriptsize SCI(5.50)}
    \end{minipage}%
}
  \subfigure{
    \begin{minipage}[t]{0.15\linewidth}
	\centering
	\includegraphics[width=1.03\linewidth]{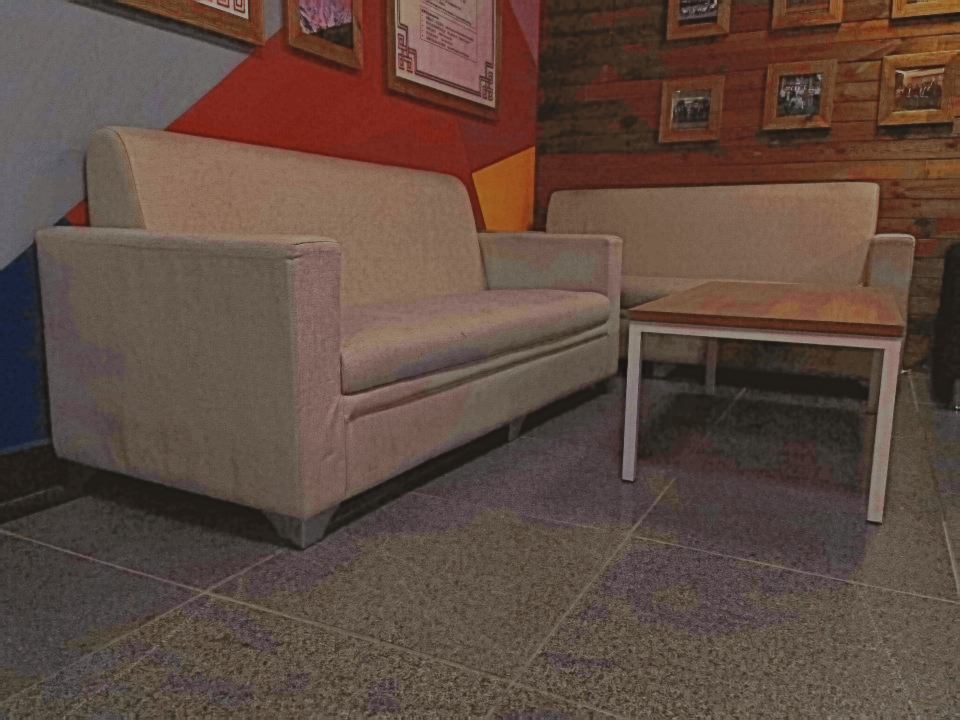}\\
        \centerline{\scriptsize PairLIE(19.19)}
	\includegraphics[width=1.03\linewidth]{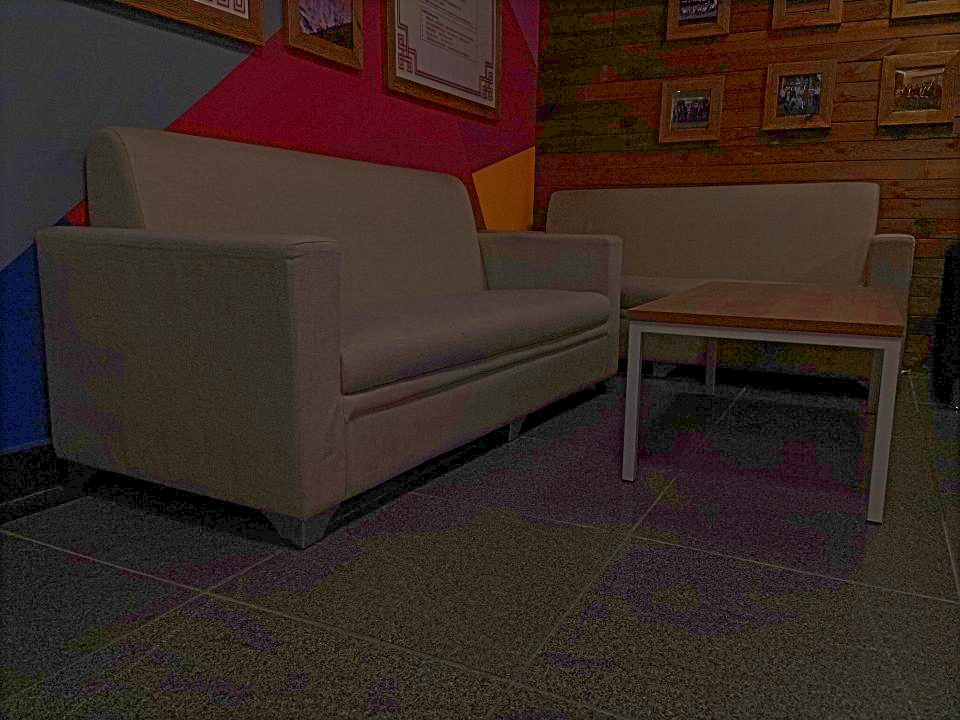}\\
        \centerline{\scriptsize CLIP-LIT(13.66)}
	\includegraphics[width=1.03\linewidth]{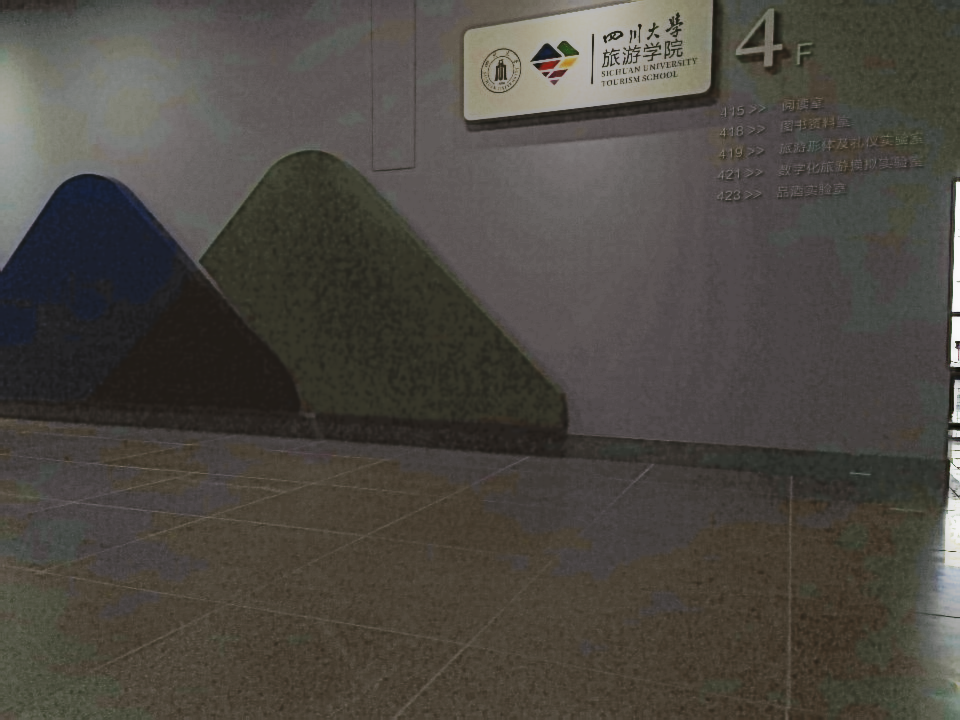}\\
        \centerline{\scriptsize PairLIE(22.27)}
	\includegraphics[width=1.03\linewidth]{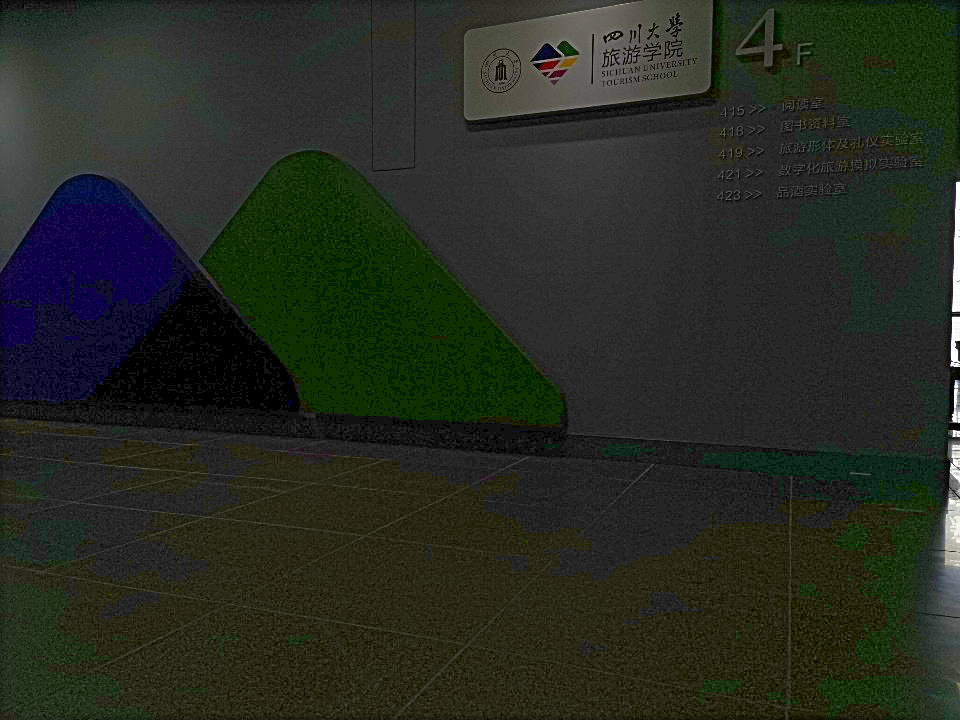}\\
        \centerline{\scriptsize CLIP-LIT(15.81)}
    \end{minipage}%
}
  \subfigure{
    \begin{minipage}[t]{0.15\linewidth}
	\centering
	\includegraphics[width=1.03\linewidth]{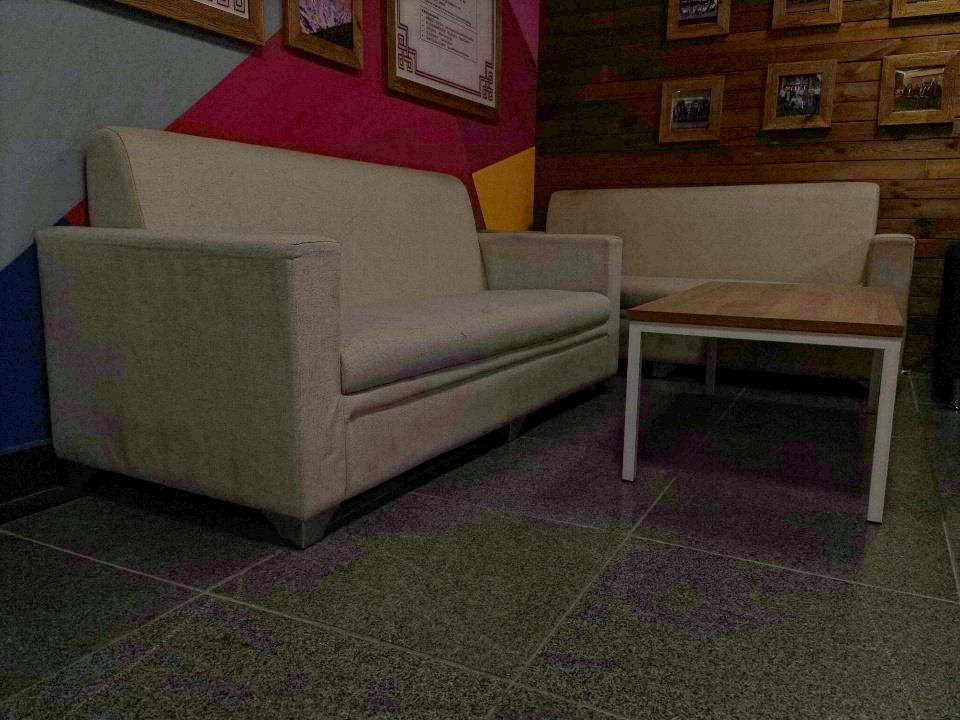}\\
        \centerline{\scriptsize ZeroDCE(15.69)}
	\includegraphics[width=1.03\linewidth]{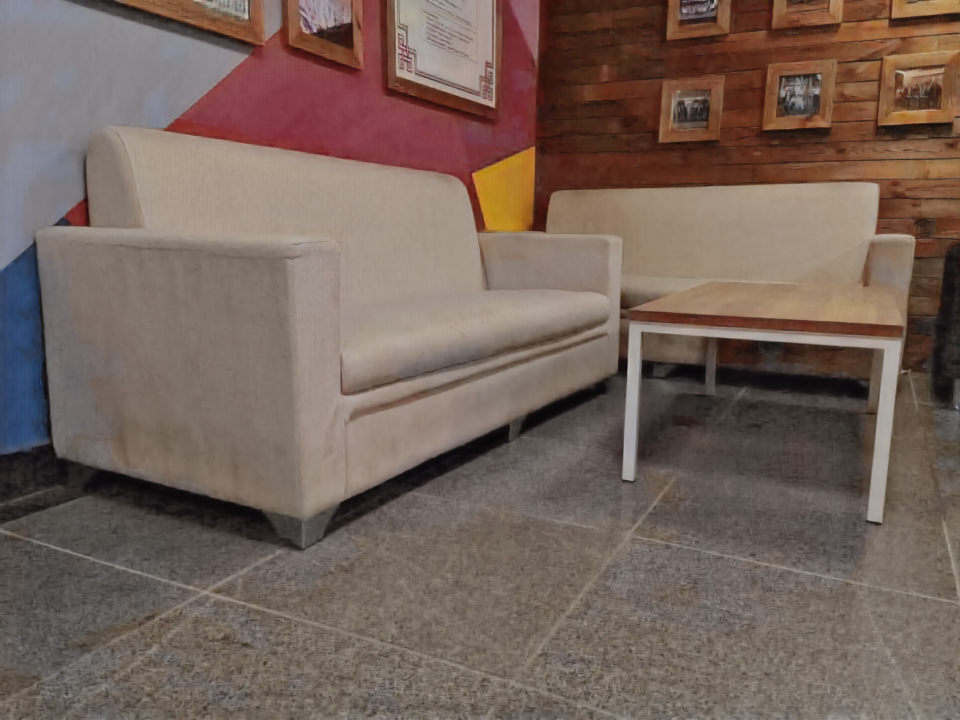}\\
        \centerline{\scriptsize Ours(18.91)}
	\includegraphics[width=1.03\linewidth]{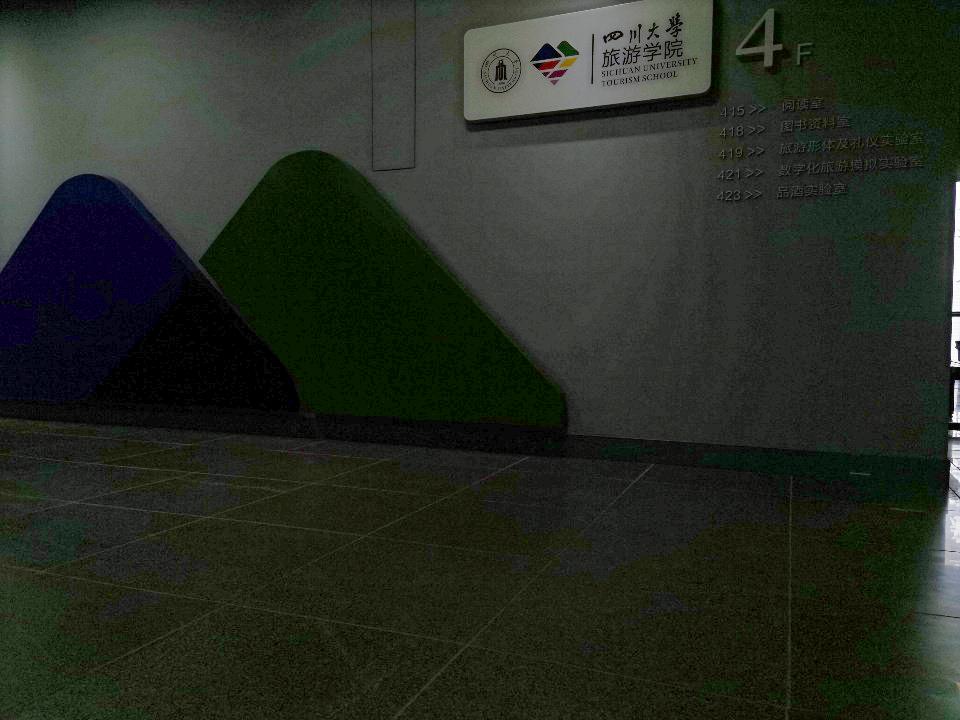}\\
        \centerline{\scriptsize ZeroDCE(17.04)}
	\includegraphics[width=1.03\linewidth]{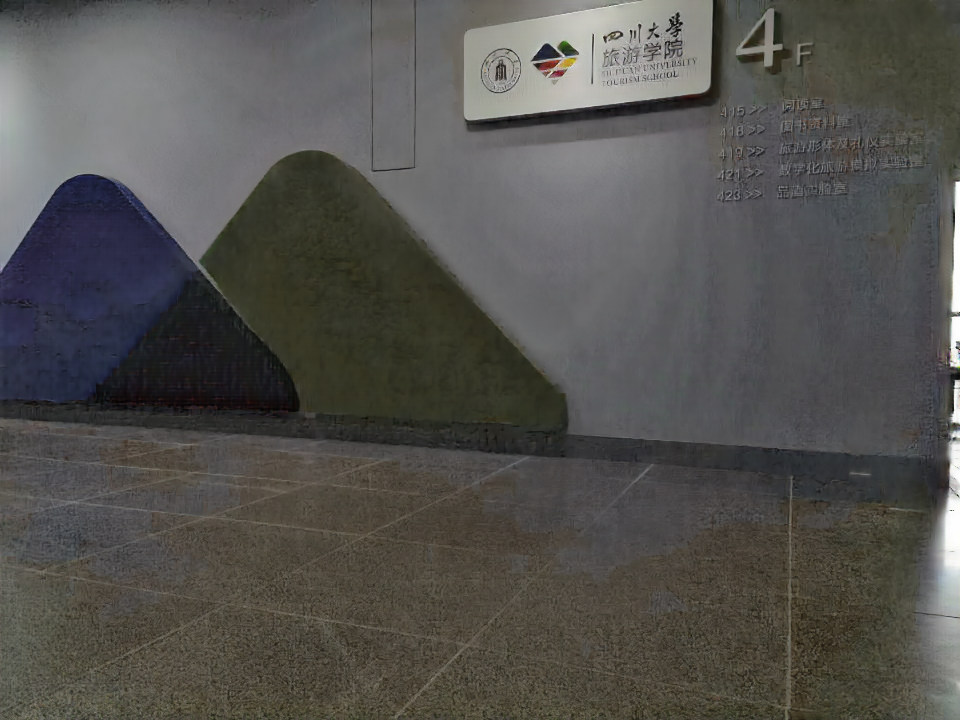}\\
        \centerline{\scriptsize Ours(22.85)}
    \end{minipage}%
} 

  \caption{Subjective comparison of the LSRW \cite{LSRW} dataset among state-of-the-art low-light image enhancement algorithms. Our method preserves more details without over-smoothing and additional artifacts. \textcolor{blue}{The corresponding PSNR values are reported below.}}
  \label{comlsrw}
\end{figure*}

\begin{figure*}[t]
  \centering
  \subfigure{
    \begin{minipage}[t]{0.15\linewidth}
	\centering
	\includegraphics[width=1.03\linewidth]{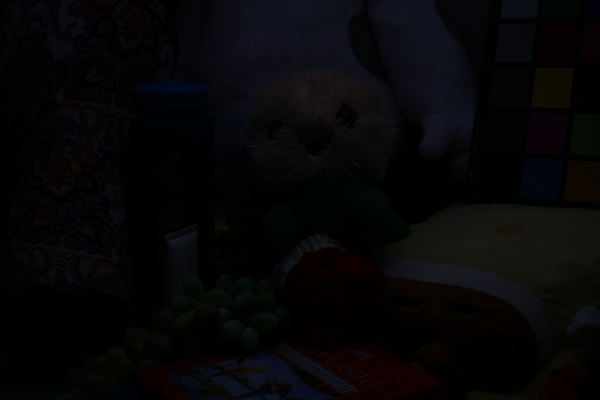}\\
        \centerline{\scriptsize Input(10.89)}
	\includegraphics[width=1.03\linewidth]{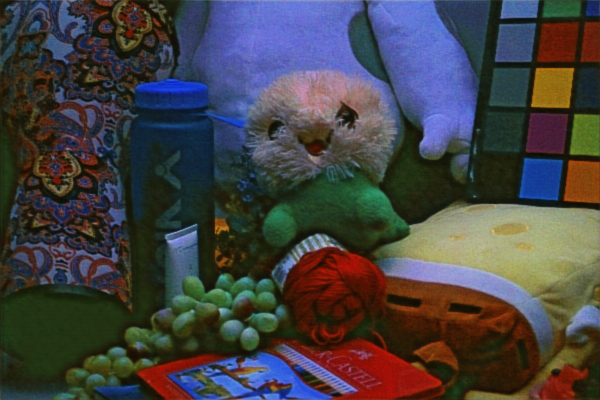}\\
        \centerline{\scriptsize SSIENet(20.63)}
    \end{minipage}%
}
  \subfigure{
    \begin{minipage}[t]{0.15\linewidth}
	\centering
	\includegraphics[width=1.03\linewidth]{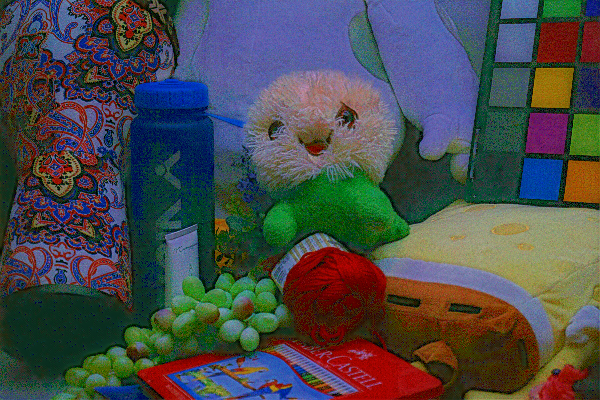}\\
        \centerline{\scriptsize RetinexNet(18.65)}
	\includegraphics[width=1.03\linewidth]{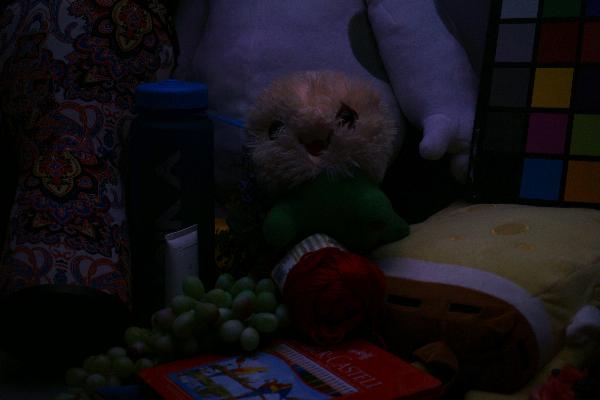}\\
        \centerline{\scriptsize RUAS(13.08)}
    \end{minipage}%
}
  \subfigure{
    \begin{minipage}[t]{0.15\linewidth}
	\centering
	\includegraphics[width=1.03\linewidth]{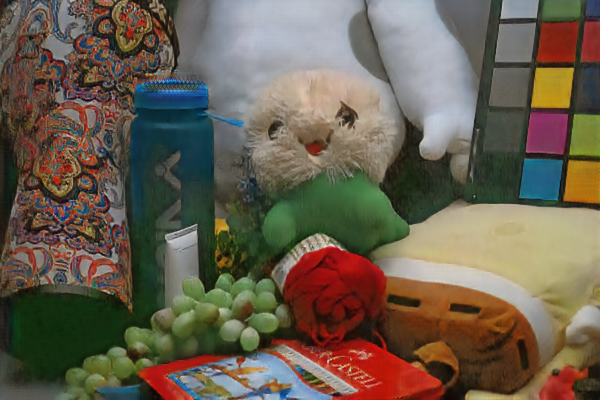}\\
        \centerline{\scriptsize KinD(16.96)}
	\includegraphics[width=1.03\linewidth]{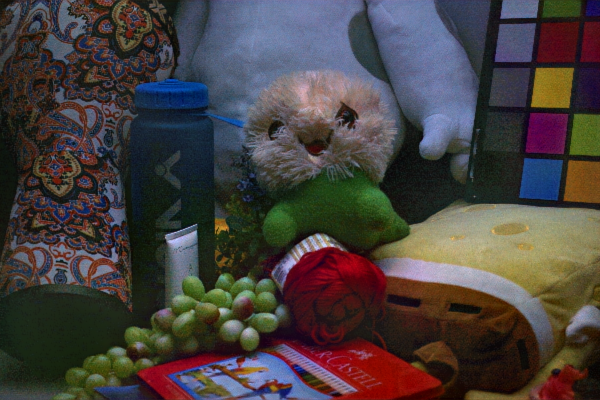}\\
        \centerline{\scriptsize EnGAN(15.55)}
    \end{minipage}%
}
  \subfigure{
    \begin{minipage}[t]{0.15\linewidth}
	\centering
	\includegraphics[width=1.03\linewidth]{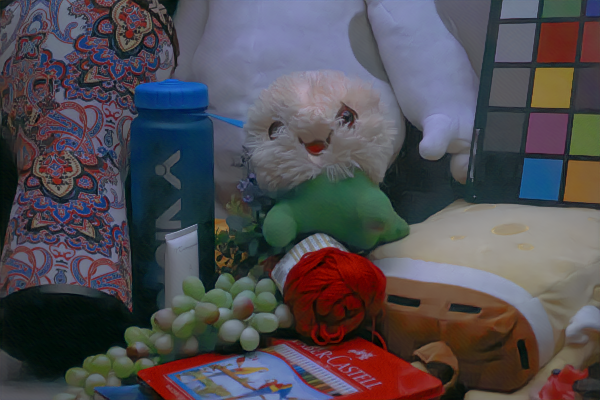}\\
        \centerline{\scriptsize URetinex(19.44)}
	\includegraphics[width=1.03\linewidth]{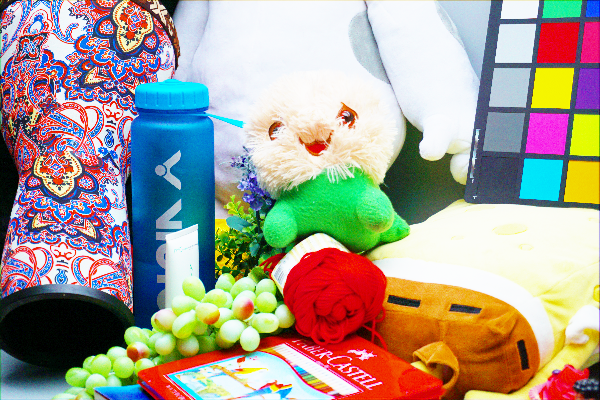}\\
        \centerline{\scriptsize SCI(7.58)}
    \end{minipage}%
}
  \subfigure{
    \begin{minipage}[t]{0.15\linewidth}
	\centering
	\includegraphics[width=1.03\linewidth]{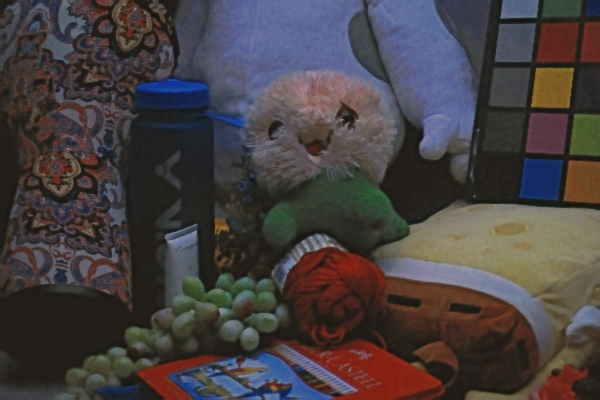}\\
        \centerline{\scriptsize PairLIE(20.17)}
	\includegraphics[width=1.03\linewidth]{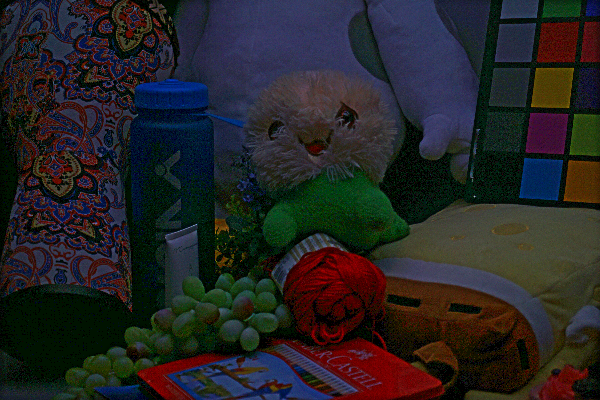}\\
        \centerline{\scriptsize CLIP-LIT(15.83)}
    \end{minipage}%
}
  \subfigure{
    \begin{minipage}[t]{0.15\linewidth}
	\centering
	\includegraphics[width=1.03\linewidth]{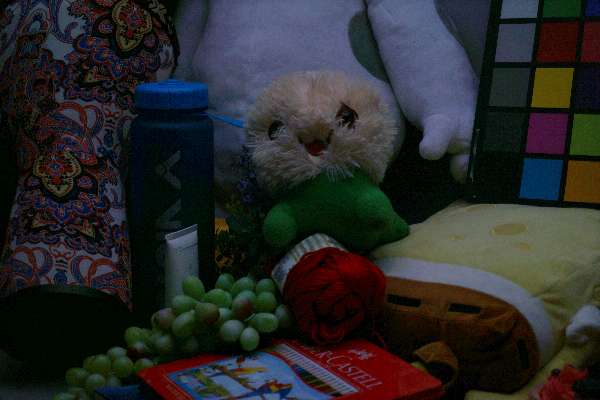}\\
        \centerline{\scriptsize ZeroDCE(17.99)}
	\includegraphics[width=1.03\linewidth]{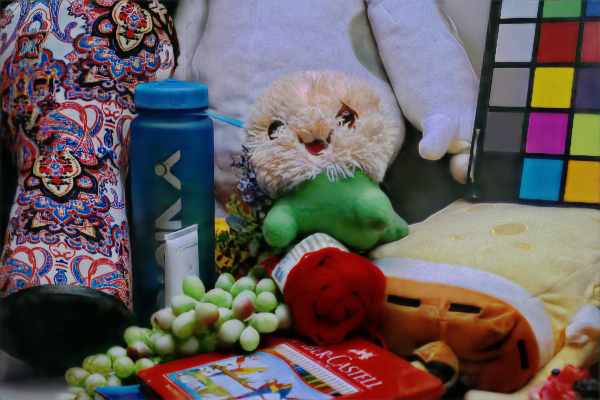}\\
        \centerline{\scriptsize Ours(21.90)}
    \end{minipage}%
} 

  \caption{Subjective comparison of the LOL \cite{LOL} dataset among state-of-the-art low-light image enhancement algorithms. Our method effectively recovers authentic lightness. \textcolor{blue}{The corresponding PSNR values are given below.}}
  \label{comlol}
\end{figure*}

\begin{table*}[t]
    \caption{\label{tab:MEF} Quantitative comparison between our method and different state-of-the-art methods on three well-known real-world benchmarks. The best and the second-best results are highlighted in \textbf{\color{red}red} and \textbf{\color{blue}blue} respectively.}
    
    \centering
    \resizebox{1.\linewidth}{!}{
    \begin{tabular}{c|c|ccc|ccccccccc}
\shline
	\multirow{2}{*}{Datasets} & \multirow{2}{*}{Metrics} & \multicolumn{3}{c|}{\cellcolor{gray!40}{Supervised Learning Methods}} & \multicolumn{8}{c}{\cellcolor{gray!40}{Unsupervised Learning Methods}} \\
	& & \cellcolor{gray!40}RetinexNet & \cellcolor{gray!40}KinD & \cellcolor{gray!40}URetinexNet & \cellcolor{gray!40}ZeroDCE & \cellcolor{gray!40}SSIENet & \cellcolor{gray!40}RUAS & \cellcolor{gray!40}EnGAN & \cellcolor{gray!40}SCI & \cellcolor{gray!40}PairLIE & \cellcolor{gray!40}CLIP-LIT & \cellcolor{gray!40}Ours \\ \shline
 	\multirow{2}{*}{DICM \cite{DICM}} & NIQE $\downarrow$ & 4.32 & \textbf{\color{red}3.34} & \textbf{\color{blue}3.43} & 3.61 & 4.04 & 4.97 & 3.57 & 3.76 & 3.54 & 3.72 & 3.51 \\
 	 & LOE $\downarrow$ & 469.2 & 259.6 & 234.2 & \textbf{\color{blue}216.5} & 532.0 & 433.3 & 403.0 & 282.7 & 287.0 & 264.6 & \textbf{\color{red}215.6} \\ \midrule
	\multirow{2}{*}{MEF \cite{MEF}} & NIQE $\downarrow$ & 4.90 & 3.38 & 3.32 & 3.31 & 4.08 & 4.11 & 3.23 & \textbf{\color{blue}3.10} & 3.92 & 3.65 & \textbf{\color{red}2.99} \\
 	 & LOE $\downarrow$ & 630.6 & 225.7 & 203.0 & 201.0 & 244.8 & 314.3 & 348.6 & \textbf{\color{red}126.8} & 218.6 & 199.9 & \textbf{\color{blue}154.1} \\ \midrule
	\multirow{2}{*}{NPE \cite{Wang_2013_TIP}} & NIQE $\downarrow$ & 4.08 & 3.53 & 4.05 & 3.48 & 3.99 & 6.23 & 4.11 & 4.07 & \textbf{\color{blue}3.49}  & 3.74& \textbf{\color{red}3.25} \\
 	 & LOE $\downarrow$ & 457.0 & \textbf{\color{red}162.4} & 225.1& 262.2 & 493.7 & 496.8 & 537.7 & 343.2 & 227.4 & 233.1 & \textbf{\color{blue}197.6} \\ 
\shline
    \end{tabular}
}
\end{table*}

\begin{figure*}[t]
  \centering
  \subfigure{
    \begin{minipage}[t]{0.15\linewidth}
	\centering
	\includegraphics[width=1.03\linewidth]{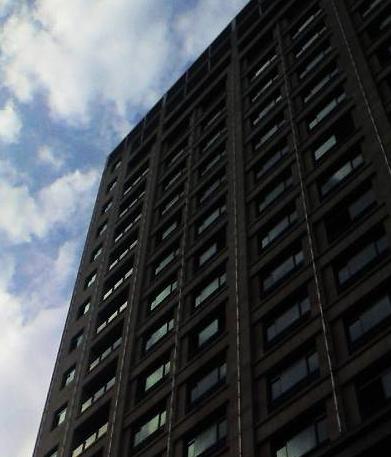}\\
        \centerline{\scriptsize Input(4.49)}
	\includegraphics[width=1.03\linewidth]{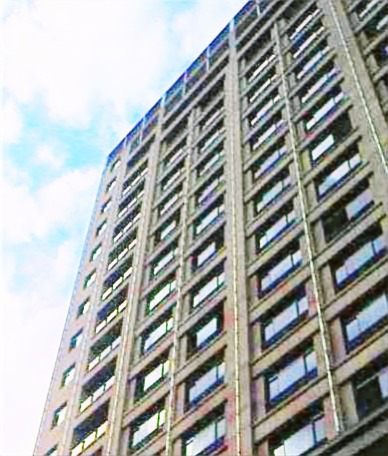}\\
        \centerline{\scriptsize SSIENet(4.73)}
    \end{minipage}%
}
  \subfigure{
    \begin{minipage}[t]{0.15\linewidth}
	\centering
	\includegraphics[width=1.03\linewidth]{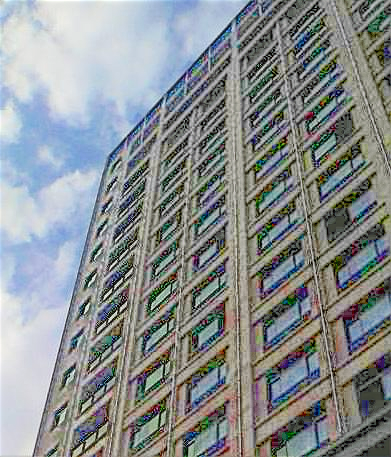}\\
        \centerline{\scriptsize RetinexNet(4.22)}
	\includegraphics[width=1.03\linewidth]{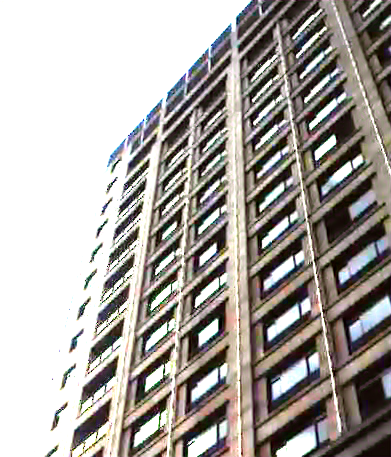}\\
        \centerline{\scriptsize RUAS(8.12)}
    \end{minipage}%
}
  \subfigure{
    \begin{minipage}[t]{0.15\linewidth}
	\centering
	\includegraphics[width=1.03\linewidth]{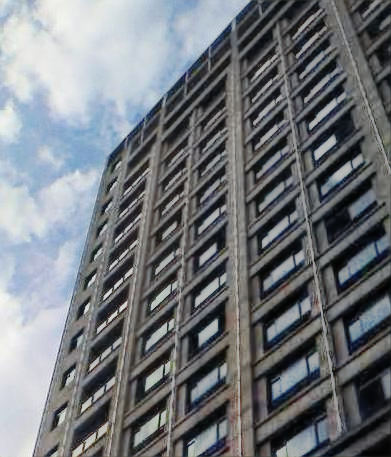}\\
        \centerline{\scriptsize KinD(4.32)}
	\includegraphics[width=1.03\linewidth]{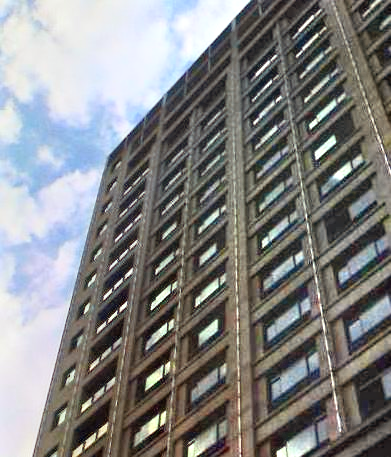}\\
        \centerline{\scriptsize EnGAN(5.29)}
    \end{minipage}%
}
  \subfigure{
    \begin{minipage}[t]{0.15\linewidth}
	\centering
	\includegraphics[width=1.03\linewidth]{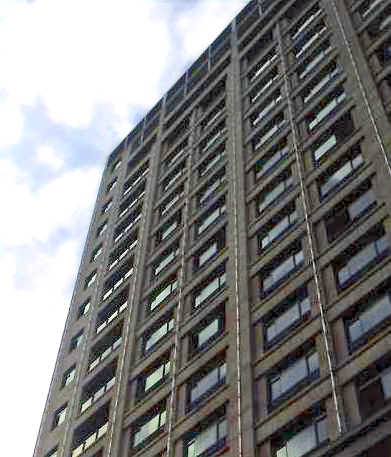}\\
        \centerline{\scriptsize URetinex(5.83)}
	\includegraphics[width=1.03\linewidth]{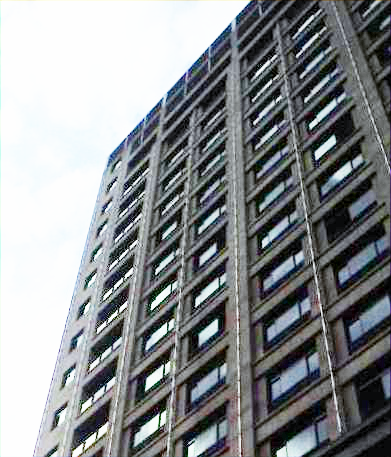}\\
        \centerline{\scriptsize SCI(5.61)}
    \end{minipage}%
}
  \subfigure{
    \begin{minipage}[t]{0.15\linewidth}
	\centering
	\includegraphics[width=1.03\linewidth]{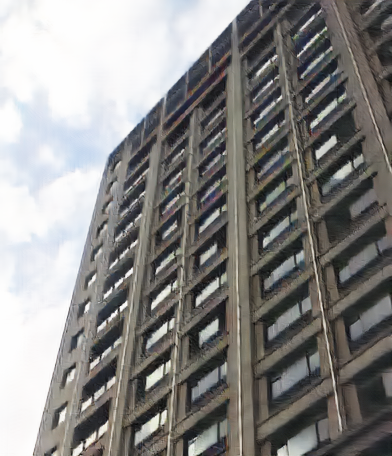}\\
        \centerline{\scriptsize PairLIE(4.60)}
	\includegraphics[width=1.03\linewidth]{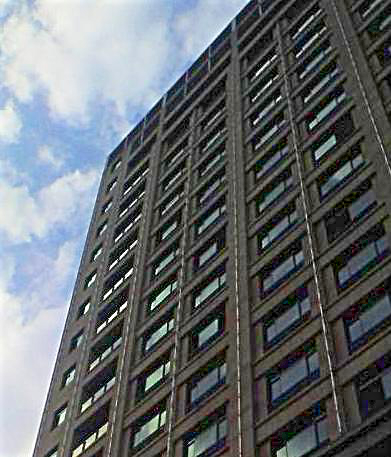}\\
        \centerline{\scriptsize CLIP-LIT(4.23)}
    \end{minipage}%
}
  \subfigure{
    \begin{minipage}[t]{0.15\linewidth}
	\centering
	\includegraphics[width=1.03\linewidth]{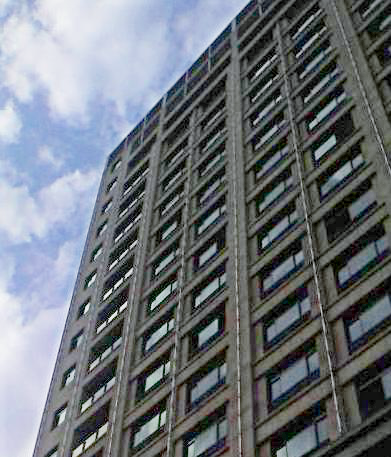}\\
        \centerline{\scriptsize ZeroDCE(4.35)}
	\includegraphics[width=1.03\linewidth]{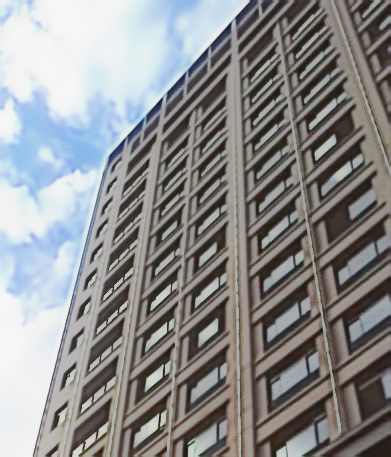}\\
        \centerline{\scriptsize Ours(4.13)}
    \end{minipage}%
} 

  \caption{Subjective comparison of the NPE \cite{Wang_2013_TIP} dataset among state-of-the-art low-light image enhancement algorithms. \textcolor{blue}{The corresponding NIQE values are given below.} One can see that our method achieves both the best visual effect and the best quantitative performance.}
  \label{comdicm}
\end{figure*}

\begin{figure*}[t]
  \centering
  \subfigure{
    \begin{minipage}[t]{0.105\linewidth}
	\centering
	\vspace{-0.8em}
        \centerline{\scriptsize Input}
	\includegraphics[width=1.05\linewidth]{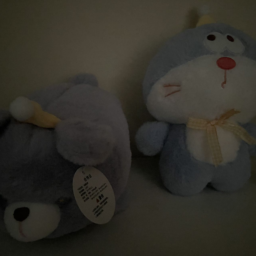}\\
        \centerline{\scriptsize 5.53}
        \includegraphics[width=1.05\linewidth]{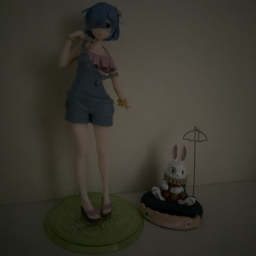}\\
        \centerline{\scriptsize 6.61}
    \end{minipage}%
}
  \subfigure{
    \begin{minipage}[t]{0.105\linewidth}
	\centering
        \centerline{\scriptsize KinD}
	\includegraphics[width=1.05\linewidth]{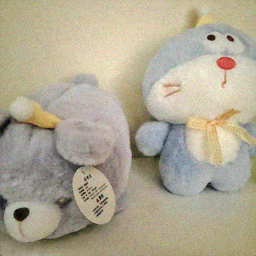}\\
        \centerline{\scriptsize 7.62}
        \includegraphics[width=1.05\linewidth]{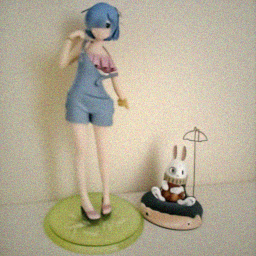}\\
        \centerline{\scriptsize 13.08}
    \end{minipage}
}
  \subfigure{
    \begin{minipage}[t]{0.105\linewidth}
	\centering
        \centerline{\scriptsize URetinex}
	\includegraphics[width=1.05\linewidth]{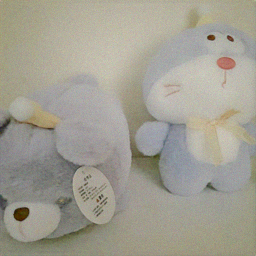}\\
        \centerline{\scriptsize 8.50}
        \includegraphics[width=1.05\linewidth]{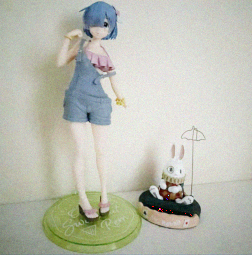}\\
        \centerline{\scriptsize 10.41}
    \end{minipage}%
}
  \subfigure{
    \begin{minipage}[t]{0.105\linewidth}
	\centering
        \centerline{\scriptsize SSIE}
	\includegraphics[width=1.05\linewidth]{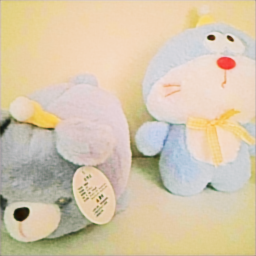}\\
        \centerline{\scriptsize 5.00}
        \includegraphics[width=1.05\linewidth]{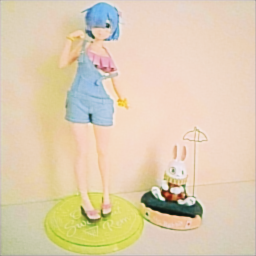}\\
        \centerline{\scriptsize 5.71}
    \end{minipage}
} 
  \subfigure{
    \begin{minipage}[t]{0.105\linewidth}
	\centering
        \centerline{\scriptsize RUAS}
	\includegraphics[width=1.05\linewidth]{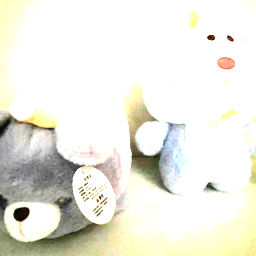}\\
        \centerline{\scriptsize 8.61}
        \includegraphics[width=1.05\linewidth]{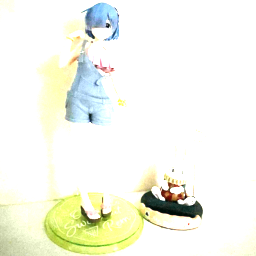}\\
        \centerline{\scriptsize 10.83}
    \end{minipage}
}
  \subfigure{
    \begin{minipage}[t]{0.105\linewidth}
	\centering
        \centerline{\scriptsize EnGAN}
	\includegraphics[width=1.05\linewidth]{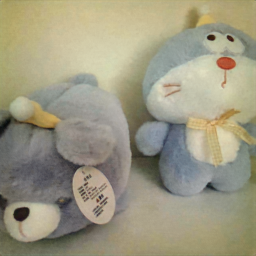}\\
        \centerline{\scriptsize 7.76}
        \includegraphics[width=1.05\linewidth]{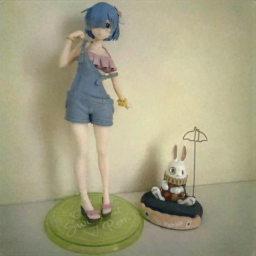}\\
        \centerline{\scriptsize 6.42}
    \end{minipage}
}
  \subfigure{
    \begin{minipage}[t]{0.105\linewidth}
	\centering
        \centerline{\scriptsize SCI}
	\includegraphics[width=1.05\linewidth]{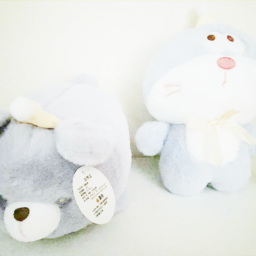}\\
        \centerline{\scriptsize 4.24}
        \includegraphics[width=1.05\linewidth]{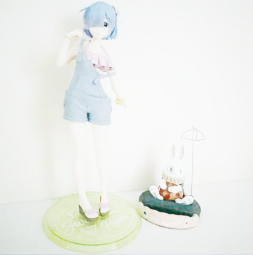}\\
        \centerline{\scriptsize 7.30}
    \end{minipage}
}
\subfigure{
    \begin{minipage}[t]{0.105\linewidth}
	\centering
        \centerline{\scriptsize Ours}
	\includegraphics[width=1.05\linewidth]{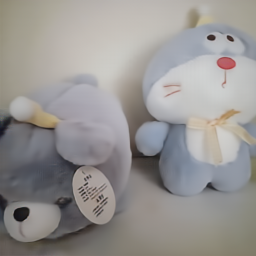}\\
        \centerline{\scriptsize 3.58}
        \includegraphics[width=1.05\linewidth]{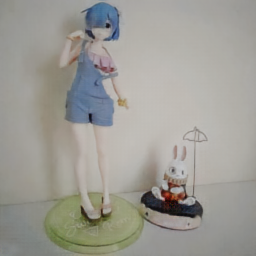}\\
        \centerline{\scriptsize 4.88}
    \end{minipage}
}
  \caption{Subjective comparison of our captured images. One can see that KinD, SSIE, and EnGAN exhibit severe color deviation while remaining obvious noise. RUAS and SCI over-expose images. URetinexNet still cannot remove noise. Whereas our method not only denoises well, but also recovers the most authentic tones. \textcolor{blue}{The corresponding NIQE values are given below.}}
  \label{fig:comreal}
\end{figure*}

\textbf{Benchmarks and Metrics.} To demonstrate the effectiveness of our method intuitively, we train and test our DiffLLE on the LSRW \cite{LSRW} and LOL \cite{LOL} datasets respectively, and additionally evaluate on the LIME \cite{Guo_2017_TIP} dataset. LSRW contains 1000 pairs of images for learning and 50 pairs for evaluation. LOL includes 485 pairs for training and 15 pairs for testing. Each pair consists of an elaborated dark image and a well-exposed reference of the same scene. Noting that during training, to prove the superiority of our unsupervised mechanism, we only adopt the low-light part of the training data and replace the normal-light references with 300 images from BSD300 dataset \cite{Martin_2001_ICCV}. LIME only contains real-world low-light images without the corresponding references. In the following experiments, we only use the LIME to test the model trained on the LSRW dataset to verify our degradation calibration capability. For the measure assessments, we use Peak Signal-to-Noise Ratio (PSNR) and Structural Similarity Index (SSIM) \cite{SSIM} to evaluate the similarity between enhanced results and the ground truth. Higher values mean more authentic results. In addition, we adopt two no-reference metrics, \textit{i.e.}, Natural Image Quality Evaluator (NIQE) \cite{Mittal_2013_SPL} and LOE \cite{Wang_2013_TIP}, to assess the quality of results. In general, a lower NIQE or LOW indicates better enhancement.

\subsection{Comparison with the State-of-the-Art Methods}
\label{sec:5.2}
We compare DiffLLE with ten State-of-the-Art (SOTA) methods. They are three well-known supervised learning methods, \textit{i.e.}, RetinexNet (BMVC 2018) \cite{LOL}, KinD (ACM MM 2019) \cite{Zhang_2019_MM}, and URetinex-Net (CVPR 2022) \cite{Wu_2022_CVPR}, and seven unsupervised methods, including ZeroDCE (CVPR 2020) \cite{Guo_2020_CVPR}, SSIENet (arxiV 2020) \cite{SSIENet}, RUAS (CVPR 2021) \cite{Liu_2021_CVPR}, EnGAN (TIP 2021) \cite{Jiang_2021_TIP}, SCI (CVPR 2022) \cite{Ma_2022_CVPR}, PairLIE (CVPR 2023) \cite{Fu_2023_CVPR}, and CLIP-LIT (ICCV 2023) \cite{liang_2023_arxiv}. Experiments are conducted on two widely used paired benchmarks (LOL \cite{LOL} and LSRW \cite{LSRW}), three well-known real-world datasets (DICM \cite{DICM}, MEF \cite{MEF}, and NPE \cite{Wang_2013_TIP}), and our captured low-light images. Since LOL and LSRW are well-crafted and their test sets are close to their training domain, there is no need to employ DDC. Hence, we only apply FTD for these benchmarks and employ complete DiffLLE (\textit{i.e.}, contains both DDC and FTD) for other three datasets and our captured real data. Here we first report results on LOL and LSRW.

\textbf{Quantitative results.} We report the quantitative comparison between our method and other SOTA methods based on two full-reference metrics (PSNR and SSIM) and two no-reference metrics (NIQE and LOE). We obtain the results of other methods by downloading their public pre-trained weights and running their official codes. As shown in Tab.~\ref{tab:compare}, compared with other unsupervised methods, our method achieves competitive results across all metrics on these two benchmarks, which are even better than some of the supervised ones. Noting that different from recent SOTA approaches (\textit{e.g.}, EnGAN \cite{Jiang_2021_TIP}, SCI \cite{Ma_2022_CVPR}, and PairLIE \cite{Fu_2023_CVPR}) that heavily rely on elaborate model structures or objective functions, our method achieves SOTA performance by only applying FTD to a simple network \cite{CycleGAN}, which validates the effectiveness of our method.

\textbf{Qualitative results.} We provide visual results on the LSRW and LOL dataset for a more intuitive illustration. As shown in Fig.~\ref{comlsrw}, the two input images from LSRW are severely degraded, whose content is barely visible. ZeroDCE, RUAS, and CLIP-LIT cannot recover enough brightness. Whilst other approaches either over-smooth background or introduce obvious veils. In contrast, our method enhances the best perceptual quality, which contains the most visual-friendly color and authentic details. The results of LOL dataset are given in Fig.~\ref{comlol}. Again, ZeroDCE and CLIP-LIT cannot enhance lightness well, whilst other methods tend to remain undesired veils. Although URetinexNet produces a clean output, it reduces the tonal contrast. Benefiting from the effective priors from the pre-trained diffusion model, the proposed FTD refines unsatisfactory coarse results and distills a high-quality solution space, achieving the best visual-friendly results.

\textbf{Extension to real-world datasets.} To demonstrate our superior generalization capabilities for out-of-domain data, we further conduct evaluation on three widely used real-world datasets, \textit{i.e.}, DICM, MEF, and NPE. In this case, we adopt complete DiffLLE, which contains the proposed DDC module to calibrate the unknown low-light degradation. Note that for the $\gamma$-curve of the DDC, we set $\gamma = 1.7$ here and it can be adjusted according to the specific input. As shown in Tab.~\ref{tab:MEF}, we compare our method with other SOTA methods and calculate two no-reference metrics (NIQE and LOE) of their results. Our method still realizes SOTA performance across all benchmarks. Noting that we directly adopt UEM trained on the LSRW dataset to enhance these images without any fine-tuning or retraining, which proves the impressive plug-and-play property of the proposed DDC module. We further provide visual results in Fig.~\ref{comdicm} for more intuitive comparison. One can see that the input image is a dark real-world scene from the NPE dataset, which is interfered by sophisticated degradation, including weak lightness and random natural noise, \textit{etc.} Other methods either cannot eliminate noise well or overexpose the image. In contrast, our method recovers the most authentic color and visual-friendly contents without any noise. We attribute this to the strong denoising ability and effective natural scene priors of the pre-trained diffusion model. The former removes unpredictable noise degradation inherent in dark regions of real scenes, and the latter helps restore more realistic content, even if they have never been included in the training data.

\textbf{Application in captured images.} Furthermore, we conduct experiments on some wild low-light images captured by ourselves to evaluate the effectiveness of different methods in practical application. As shown in Fig.~\ref{fig:comreal}, we compare our method with SOTA algorithms proposed in recent years. One can see that although some approaches (such as KinD, SSIE and EnGAN) recover enough lightness, they severely distort the hue and still remain obvious natural noise. RUAS and SCI tend to over-expose images and URetinexNet cannot denoise well either. On the contrary, our method recovers the most realistic color while removing almost all noise. We attribute it to the effective denoising capacity and prolific natural priors of the diffusion model, which calibrates complicated interferences to decrease enhancement difficulty.

\subsection{Ablation Study}
\label{sec:5.3}
Here we discuss the effect of each proposed component in detail, which can be calssified to three settings. \textbf{i)} ``\#1'' is a naive CycleGAN without any other operations. \textbf{ii)} ``\#2'' is the setting with the developed Fin-grained Target domain Distillation (FTD) operation. \textbf{iii)} Finally, we add the proposed Diffusion-guided Degradation Calibration (DDC) to complete DiffLLE. Experiments are conducted on both in-domain and out-of-domain data.

\textbf{In-domain data.} In Tab.~\ref{tab:ID}, we train the model w/ and w/o the proposed FTD operation on the LSRW and LOL datasets, and then evaluate the trained model on the test set of the corresponding dataset. Since the training data and test data belong to the same benchmark, the test set can be regarded as the in-domain data. Hence, there is no need to use degradation calibration and here we only validate the effect of FTD. Apparently, since the lack of explicit strong supervision, this unsupervised model cannot constrain its learned normal-light target space to a fine enough high-quality field. As shown in the upper row of Tab.~\ref{tab:ID}, in both sets of experiments, the naive model only achieves limited quality enhancement in the coarse normal-light domain. In contrast, our FTD re-calibrates the target domain and achieves impressive improvements on both datasets, as shown in the lower row of Tab.~\ref{tab:ID}. It demonstrates that FTD distills a higher quality fine solution from the coarse brightness domain, resulting in impressive performance gains.

\textbf{Out-of-domain data.} Since learning-based methods heavily rely on training data, it is challenging to enhance degradation features that are not included in the training set. Here we evaluate our method on the out-of-domain data to validate its superior generalization. Specifically, we train all settings on the LSRW dataset and test them on other datasets. As the training data and test data come from different benchmarks, the test set can be regarded as out-of-domain data. As shown in Tab.~\ref{tab:OOD}, LOL is a paired low-light-normal-light dataset, and LIME only contains low-light images without the ground truth. One can see that the naive setting \#1 only recovers a coarse solution with the worst quantitative values. Benefiting from the fine-grained domain distilled by FTD, performance has been improved. But the inherent gap between domains restricts the effectiveness. To this end, the proposed DDC calibrates the input low-light image in the feature space, which narrows the gap between the input and the learned degradation domain. Taking the results on the LOL dataset as an example, \#2 improves PSNR by 2.1$\%$, whilst DiffLLE further improves PSNR by 7.2$\%$ based on \#2. It proves that DDC has unique advantages in enhancing out-of-domain data. In addition, we capture some real-world dark images ourselves to validate the capability of DDC in out-of-domain cases. Results are given in Sec.~\ref{sec:5.4}.

\begin{table}
  \caption{\label{tab:ID} Ablation study on the proposed Fin-grained Target domain Distillation (FTD). The best and the second-best results are highlighted in \textbf{\color{red}red} and \textbf{\color{blue}blue} respectively.}
  
  \centering
  \resizebox{1.\linewidth}{!}{
  \begin{tabular}{c|c|c|cccc}
    \shline
    \cellcolor{gray!40}Case &\cellcolor{gray!40}Test Set & \cellcolor{gray!40}FTD & \cellcolor{gray!40}PSNR $\uparrow$ & \cellcolor{gray!40}SSIM $\uparrow$ & \cellcolor{gray!40}NIQE $\downarrow$ & \cellcolor{gray!40}LOE $\downarrow$ \\ \shline
    \#1 & \multirow{2}{*}{LSRW} & $\times$ & \textbf{\color{blue}16.77} & \textbf{\color{blue}0.4565} & \textbf{\color{blue}3.32} & \textbf{\color{blue}272.4} \\
    \#2 &  & \checkmark & \textbf{\color{red}18.63} & \textbf{\color{red}0.5536} & \textbf{\color{red}2.79} & \textbf{\color{red}201.3} \\ \midrule
    \#1 & \multirow{2}{*}{LOL} & $\times$ & \textbf{\color{blue}16.42} & \textbf{\color{blue}0.5784} & \textbf{\color{blue}4.97} & \textbf{\color{blue}249.0} \\
    \#2 &  & $\checkmark$ & \textbf{\color{red}22.24} & \textbf{\color{red}0.7923} & \textbf{\color{red}3.09} & \textbf{\color{red}202.4} \\
    \shline
  \end{tabular}
}
\end{table}

\begin{table}[t]
    \caption{\label{tab:OOD} Ablation study on three settings. The best and the second-best results are highlighted in \textbf{\color{red}red} and \textbf{\color{blue}blue} respectively. All settings are only trained on the LSRW dataset.}
    
    \centering
    \resizebox{1.\linewidth}{!}{
    \begin{tabular}{c|c|cc|cccc}
\shline
	\cellcolor{gray!40}Case &\cellcolor{gray!40}Test Set & \cellcolor{gray!40}DDC & \cellcolor{gray!40}FTD & \cellcolor{gray!40}PSNR $\uparrow$ & \cellcolor{gray!40}SSIM $\uparrow$ & \cellcolor{gray!40}NIQE $\downarrow$ & \cellcolor{gray!40}LOE $\downarrow$ \\ \shline
	\#1 & \multirow{3}{*}{LOL} & $\times$ & $\times$ & 16.89 & 0.6854 & 4.93 & 274.5  \\
 	\#2 &  & $\times$ & $\checkmark$ & \textbf{\color{blue}17.25} & \textbf{\color{blue}0.7491} & \textbf{\color{blue}4.25} & \textbf{\color{blue}246.1} \\
	DiffLLE &  & $\checkmark$ & $\checkmark$ & \textbf{\color{red}18.50} & \textbf{\color{red}0.7523} & \textbf{\color{red}3.42} & \textbf{\color{red}236.8}       \\ \midrule
        \#1 & \multirow{3}{*}{LIME} & $\times$ & $\times$ & - & - & 4.90 & 261.0  \\
        \#2 &  & $\times$ & $\checkmark$ & - & - & \textbf{\color{blue}4.17} & \textbf{\color{blue}211.5} \\
        DiffLLE &  & $\checkmark$ & $\checkmark$ & - & - & \textbf{\color{red}3.76} & \textbf{\color{red}206.3} \\
\shline
    \end{tabular}
}
\end{table}

\begin{table}[t]
    \caption{\label{tab:CDS} The Cross Discrminator Score (CDS) of LSRW and LOL datasets. We compute it using discriminators trained on the LSRW and LOL datasets respectively. The best and second best results are emphasized in \textbf{\color{red}red} and \textbf{\color{blue}blue} respectively.}
    
    \centering
    \resizebox{1.\linewidth}{!}{
    \begin{tabular}{c|cc|cc}
\shline
	\cellcolor{gray!40}Testing Data & \cellcolor{gray!40}LSRW (w/o) & \cellcolor{gray!40}LSRW (w/) & \cellcolor{gray!40}LOL (w/o) & \cellcolor{gray!40}LOL (w/) \\ \hline
	CDS $\uparrow$ & \textbf{\color{blue}0.4021} & \textbf{\color{red}0.5599} & \textbf{\color{blue}0.5337} & \textbf{\color{red}0.5772} \\
\shline
    \end{tabular}
}
\end{table}

\subsection{Analysis}
\label{sec:5.4}
In this section, we first analyze the effect of the proposed Diffusion-guided Degradation Calibration (DDC) operation, and conduct experiments on our captured real-world scenario. We develop a novel metric, named CDS, to evaluate the generalization ability. Besides, we compare the computational performance of our method and other approaches. Furthermore, we also study the effect of iterations in the reverse process. Finally, we apply the proposed two plug-and-play modules, \textit{i.e.}, DDC and FTD, to other unsupervised baselines and achieve performance gains.

\textbf{Calibration analysis.} We conduct a comprehensive analysis of our Diffusion-guided Degradation Calibration (DDC) from two perspectives. First, we utilize images from the LOL dataset denoted as LOL (w/o) and the images calibrated by DDC as LOL (w/). The same nomenclature is applied to the LSRW dataset. For quantitative evaluation, we employ a trained discriminator based on CycleGAN. This discriminator produces the probabilities of which an input image conforms to the feature distribution of images in the dataset. A higher probability value indicates a more consistent distribution. Noting that for a model trained on a specific dataset, the images of other datasets are out-of-domain data. Consequently, we train the discriminator on the LSRW dataset and utilize it to evaluate LOL (w/o) and LOL (w/), and vice versa for the discriminator trained on the LOL dataset. Given that we adopt the discriminator structure of patchGAN, which outputs a map with values not strictly limited between zero and one, we calculate the mean value of the output map and normalize it for a more intuitive evaluation. This normalized probability is termed Cross Discriminator Score (CDS) in this paper, representing a novel metric for assessing the generalization ability. As depicted in Table~\ref{tab:CDS}, for both datasets, the calibrated images exhibit higher CDS values than the original images, thus demonstrating the effectiveness of our diffusion-guided degradation calibration strategy.

\begin{figure}[t]
  \centering
  \subfigure{
    \begin{minipage}[t]{0.22\linewidth}
	\centering
	\includegraphics[width=1.03\linewidth]{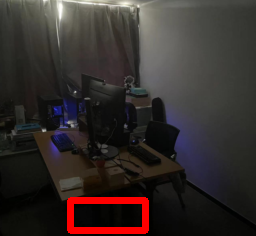}\\
        \includegraphics[width=1.03\linewidth]{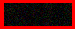}\\
        \centerline{\scriptsize Input}

    \end{minipage}%
}
  \subfigure{
    \begin{minipage}[t]{0.22\linewidth}
	\centering
	\includegraphics[width=1.03\linewidth]{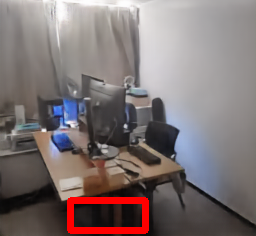}\\
        \includegraphics[width=1.03\linewidth]{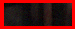}\\ 
        \centerline{\scriptsize w/o DDC}

    \end{minipage}%
}
  \subfigure{
    \begin{minipage}[t]{0.22\linewidth}
	\centering
        \includegraphics[width=1.03\linewidth]{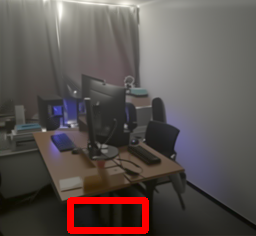}\\
        \includegraphics[width=1.03\linewidth]{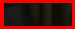}\\
        \centerline{\scriptsize DDC}
    \end{minipage}%
}
  \subfigure{
    \begin{minipage}[t]{0.22\linewidth}
	\centering
        \includegraphics[width=1.03\linewidth]{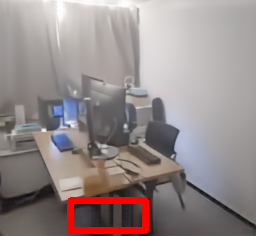}\\
        \includegraphics[width=1.03\linewidth]{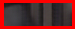}\\
        \centerline{\scriptsize Ours}
    \end{minipage}%
}
  \caption{Subjective results of two ablation settings (\#2 and Ours) and the proposed DDC module. One can see that without DDC, the enhanced result still maintains natural noise and invisible regions (such as the floor), DDC adjusts uneven brightness distribution and denoises. The complete setting recovers a more clean result with uniform brightness.}
  \label{fig:capture}
\end{figure}

Furthermore, we apply our method on real low-light images to validate the robustness of calibration. As shown in Fig.~\ref{fig:capture}, we capture a low-light image of our laboratory. It contains many objects and has a machine-generated blue glow on the wall, which is a classical sophisticated real-world low-light scene. Without DDC, we directly utilize the unsupervised model (\textit{i.e.}, \#2) to enhance it and obtain the result ``w/o DDC''. One can see that although we recovered most of the areas, some are still invisible, such as the region under the desk. Besides, the inherent glow on its wall stands out even more. In DiffLLE, we first employ DDC to pre-process input, as shown in ``DDC'' of Fig.~\ref{fig:capture}. Benefiting from the brightness curve, extreme lightness is adjusted well, such as the floor. Whilst diffusion-based domain calibration refines this coarse low-light image, eliminating its natural noise and supplementing authentic details. Then we apply an enhancement module to it to recover a high-quality normal-light scene. We display it in ``Ours'' of Fig.~\ref{fig:capture}. It not only decreases the glow on the wall but also enhances it with more uniform brightness. For example, the floor becomes visible.

\begin{table}[t]
    \caption{\label{tab:FLOP} The GFLOPS, parameter number, and running time of different methods and their performance on the LSRW dataset.}
    
    \centering
    \resizebox{1.\linewidth}{!}{
    \begin{tabular}{c|ccccc}
\shline
	\cellcolor{gray!40}Method & \cellcolor{gray!40}DiffLLE & \cellcolor{gray!40}EnGAN & \cellcolor{gray!40}URetinexNet & \cellcolor{gray!40}CLIP-LIT & \cellcolor{gray!40}PairLIE\\ \hline
	GFLOPS & 56.86 & \textbf{\color{red}18.15} & 56.93 & \textbf{\color{blue}18.21} & 22.35  \\
        Params. (M) & 11.378 & 54.410 & \textbf{\color{blue}0.340} & \textbf{\color{red}0.279} & 0.342 \\
        Time (ms) & 6.44 & 6.75 & 8.91 & \textbf{\color{red}2.28} & \textbf{\color{blue}2.30} \\
        PSNR (dB) & \textbf{\color{red}18.63} & 17.06 & \textbf{\color{blue}18.10} & 13.48 & 17.60 \\
        SSIM & \textbf{\color{red}0.5536} & 0.4601 & \textbf{\color{blue}0.5149} & 0.3962 & 0.5009 \\
\shline
    \end{tabular}
}
\end{table}

\textbf{Computational performance.} Speed and computational burden are also crucial for unsupervised methods applied in real-world scenarios. We further analyze computation burden of our method and some other SOTA methods in Tab.~\ref{tab:FLOP}, which compares from three perspectives, \textit{i.e.}, GFLOPS, parameter number, and running time. Specifically, we input ten images with a size of 256×256 to calculate GFLOPs and the average time consumption on a single NVIDIA 3090 GPU. Notably, our approach still performs competitive speed with a significant advantage in PSNR and SSIM. Although it is not the fastest method, its unparalleled enhancement quality and robustness in addressing real-world degradations have been extensively demonstrated and validated in this paper.

\begin{figure}[t]
  \centering
  \includegraphics[width=1.0\linewidth]{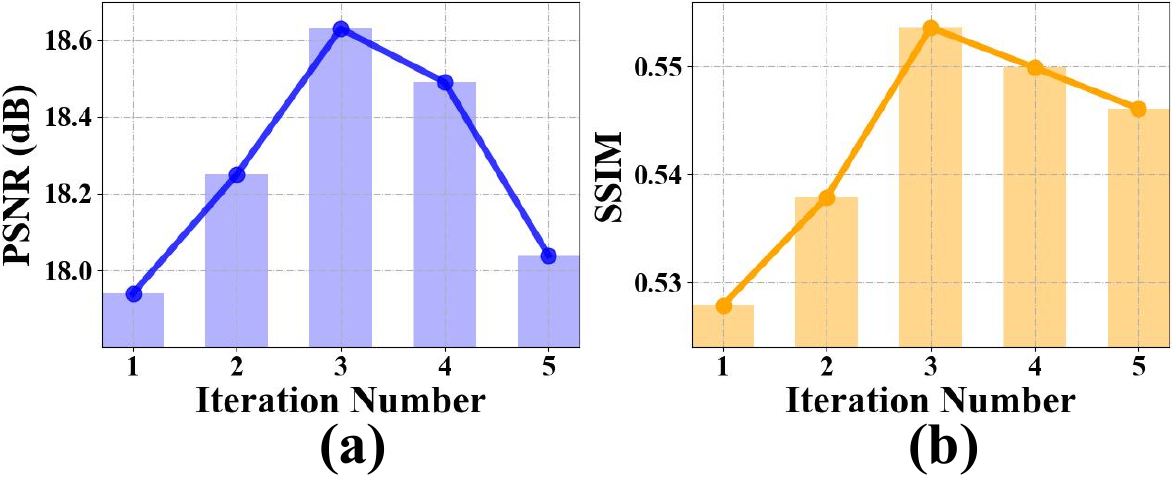}

  \caption{Ablation study on different iteration numbers. We distill the target domain with different iterations. Obviously, the best iteration number is 3.}
  \label{fig:psnr}
\end{figure}

\textbf{Iterations analysis.} We study the effect of the iteration number of adding and removing noise on model performance. Specifically, here we adjust iteration number of the FTD module for a simple illustration. As shown in Fig.~\ref{fig:psnr}, we conduct different numbers of iterations to the preliminary enhanced results to generate different versions of pseudo-ground-truths and use them to distill the target domain. Experiments are performed on the LSRW dataset. One can see that, both PSNR and SSIM first increase and then decrease, and both are the highest when the number of iterations is set to 3. We suppose this is because, on the one hand, the artifacts and chromatic aberrations in the initial coarse target domain cannot be removed well when the number of iterations is small. On the other hand, when iterating too many steps, the desired details and structures are falsified by the generative model, resulting in distortions from references. In this paper, we set the inversion number to 3 in other experiments for the trade-off between fine quality and fidelity.

\begin{table}[t]
    \caption{\label{tab:ZeroDCE} We apply the proposed DDC module on other unsupervised methods and test on three well-known real-world benchmarks. The best and second best results of each set are emphasized in \textbf{\color{red}red} and \textbf{\color{blue}blue} respectively.}
    
    \centering
    \resizebox{1.\linewidth}{!}{
    \begin{tabular}{c|c|ccc}
\shline
	\cellcolor{gray!40}Methods & \cellcolor{gray!40}Settings & \cellcolor{gray!40}DICM & \cellcolor{gray!40}MEF & \cellcolor{gray!40}NPE \\ \hline
	\multirow{2}{*}{ZeroDCE} & w/o DDC & \textbf{\color{blue}3.61} / \textbf{\color{blue}216.5} & \textbf{\color{blue}3.31} / \textbf{\color{blue}201.0} & \textbf{\color{blue}3.48} / \textbf{\color{blue}262.7} \\
        & w/ DDC & \textbf{\color{red}3.01} / \textbf{\color{red}211.0} & \textbf{\color{red}3.10} / \textbf{\color{red}177.8} & \textbf{\color{red}3.22} / \textbf{\color{red}221.4} \\ \hline
        \multirow{2}{*}{RUAS} & w/o DDC & \textbf{\color{blue}4.97} / \textbf{\color{blue}433.3} & \textbf{\color{blue}4.08} / \textbf{\color{blue}314.3} & \textbf{\color{blue}6.23} / \textbf{\color{blue}496.8} \\
        & w/ DDC & \textbf{\color{red}3.54} / \textbf{\color{red}380.4} & \textbf{\color{red}3.46} / \textbf{\color{red}281.6} & \textbf{\color{red}3.91} / \textbf{\color{red}372.3} \\ \hline
        \multirow{2}{*}{EnGAN} & w/o DDC & \textbf{\color{blue}3.57} / \textbf{\color{blue}403.0} & \textbf{\color{red}3.13} / \textbf{\color{blue}348.6} & \textbf{\color{blue}4.11} / \textbf{\color{blue}537.7} \\
        & w/ DDC & \textbf{\color{red}3.26} / \textbf{\color{red}305.1} & \textbf{\color{blue}3.34} / \textbf{\color{red}254.5} & \textbf{\color{red}3.47} / \textbf{\color{red}327.9} \\
\shline
    \end{tabular}
}
\end{table}

\begin{table}[t]
    \footnotesize
    \caption{\label{tab:EGAN} We apply the proposed FTD module to other unsupervised methods and test on two well-known paired benchmarks. The best and second best results of each set are emphasized in \textbf{\color{red}red} and \textbf{\color{blue}blue} respectively.}
    
    \centering
    \resizebox{1.\linewidth}{!}{
    \begin{tabular}{c|c|cc}
    \shline
    \cellcolor{gray!40}Methods & \cellcolor{gray!40}Settings & \cellcolor{gray!40}LOL & \cellcolor{gray!40}LSRW \\ \hline
    
    \multirow{2}{*}{EnGAN} & w/o FTD & \textbf{\color{red}18.13} / \textbf{\color{blue}0.6655} & \textbf{\color{blue}17.06} / \textbf{\color{blue}0.4601}\\  
    & w/ FTD & \textbf{\color{blue}18.04} / \textbf{\color{red}0.7053} & \textbf{\color{red}17.49} / \textbf{\color{red}0.5307} \\  \hline
    
    \multirow{2}{*}{CLIP-LIT} & w/o FTD & \textbf{\color{blue}12.39} / \textbf{\color{blue}0.4934} & \textbf{\color{blue}13.48} / \textbf{\color{blue}0.3962} \\
    & w/ FTD & \textbf{\color{red}13.06} / \textbf{\color{red}0.5371} & \textbf{\color{red}13.75} / \textbf{\color{red}0.4505} \\  \hline
    
    \multirow{2}{*}{PairLIE} & w/o FTD & \textbf{\color{blue}19.51} / \textbf{\color{blue}0.7358} & \textbf{\color{blue}17.60} / \textbf{\color{blue}0.5009} \\
    & w/ FTD & \textbf{\color{red}20.04} / \textbf{\color{red}0.7820} & \textbf{\color{red}18.09} / \textbf{\color{red}0.5340} \\
    
    \shline
    \end{tabular}
}
\end{table}

\textbf{Application to other baselines.} To prove the plug-and-play property of the proposed DDC and FTD modules, here we apply them to other pipelines. The iteration number of FTD and DDC are also set to 3. As shown in Tab.~\ref{tab:ZeroDCE}, we replace CycleGAN with other three unsupervised methods, \textit{i.e.}, ZeroDCE, RUAS, and EnGAN, and test on three widely used real-world benchmarks without any fine-tuning or retraining. The NIQE / LOE values are displayed. One can see that although all of the unsupervised algorithms are elaborate, the proposed DDC module still realizes obvious performance gains, which proves its superior effectiveness. In addition, we also conduct alternative experiments on the FTD module to verify its distillation capacity for other unsupervised methods. As FTD aims to generate pseudo-references and fine-tune the enhancement model based on them, here we take EnGAN, CLIP-LIT, and PairLIE as baselines and test on LOL and LSRW datasets. Their PSNR / SSIM values are given in Tab.~\ref{tab:EGAN}. Although all of these methods have been able to perform well on these data, our FTD still further improves their performance, proving its impressive capacity of fine-tuning.


\section{Conclusion}
In this paper, we propose a Diffusion-based domain calibration strategy to achieve effective unsupervised Low-Light image Enhancement, dubbed DiffLLE. Specifically, we design two novel plug-and-play modules to improve generalization ability and enhancement effect. For the coarse target domain learned by an unsupervised framework, the developed Fine-grained Target domain Distillation (FTD) refines a high-quality normal-light field. Since the lack of explicit supervision, the unsupervised model inevitably introduces artifacts and unnatural tones, which severely affect visual perception. Through diffusion-based domain calibration, FTD distills a fine subset from this coarse domain to find the optimal solution. For dark degradation scenarios not covered in the training data, we propose a pre-processing technique, namely Diffusion-guided Degradation Calibration (DDC). It consists of a lightness enhancement curve and a diffusion-based domain calibration process. The former adjusts extreme brightness in real-world scenes and the latter eliminates natural noise and captured artifacts. In this way, real-world stochastic degradations are calibrated to approximate the elaborate training domain, improving our generalization. Note that both two components are plug-and-play, experiments have proved their superiority compared with other top-performing methods.

\textbf{Limitations.} Although we have proposed a SOTA method for unsupervised low-light image enhancement, there are still several limitations. First, limited by the bulky diffusion model, our method does not have an advantage in computational performance. Second, we only train our model on limited training data (LSRW and LOL), which restricts its application to real-world scenes, since many objects and phenomena have not been learned. For example, light is reflected on the smooth surface, and our method is not yet able to enhance this effect well.

\textbf{Future work.} We plan to apply existing sampling acceleration strategy to decrease our computational burden, such as consistency model \cite{Song_2023_arxiv}, token merging \cite{Bolya_2023_ICLR}, and distillation \cite{Meng_2023_CVPR}. In addition, the training data used in this paper, \textit{i.e.}, LOL and LSRW, are well-designed limited datasets. In the future, we plan to train on more complex real-world data to improve practical application performance. Meanwhile, the proposed FTD and DDC will also play a beneficial role in other low-level vision tasks such as image deraining \cite{Jiang_2023_arxiV, Liu_2020_TNNLS}, underwater imaging \cite{Ye_2022_CVPR}, hyperspectral imaging \cite{zhang2022herosnet, zhang2023progressive}, compressive sensing \cite{chen2023tip} and omnidirectional image super-resolution~\cite{sun2023opdn}.

\bibliography{sn-bibliography}

\end{document}